\documentclass[preprint,review,12pt]{elsarticle}

\usepackage{amssymb}
\usepackage{algpseudocode}
\usepackage{algorithm}
\usepackage[page,toc,titletoc,title]{appendix}

\makeatletter 
 
\@addtoreset{algorithm}{section} 
\makeatother

\usepackage{listings}
\usepackage{bm}
\usepackage{pgfplots}
\usepackage{pstricks,pst-node, pst-fun}
\usepackage{dsfont}
\usepackage{amssymb}
\usepackage[british,UKenglish]{babel}
\usepackage[utf8]{inputenc}
\usepackage[a4paper,left=3cm,right=2cm,top=2.5cm,bottom=2.5cm]{geometry}
\usepackage{amsmath}
\usepackage{relsize}
\usepackage{setspace}
\usepackage{subfig}
\usepackage[symbol]{footmisc}

\usepackage{tabularx}
\usepackage{easybmat}
\usepackage{multirow}
\usepackage{hhline}
\usepackage[hidelinks]{hyperref}

\DeclareFixedFont{\ttb}{T1}{txtt}{bx}{n}{12} 
\DeclareFixedFont{\ttm}{T1}{txtt}{m}{n}{12}  

\usepackage{color}
\definecolor{deepblue}{rgb}{0,0,0.5}
\definecolor{deepred}{rgb}{0.6,0,0}
\definecolor{deepgreen}{rgb}{0,0.5,0}

\usepackage{listings}

\newcommand\pythonstyle{\lstset{
language=Python,
basicstyle=\ttm,
morekeywords={self,False,True},              
keywordstyle=\ttb\color{deepblue},
emph={MyClass,__init__},          
emphstyle=\ttb\color{deepred},    
stringstyle=\color{deepgreen},
frame=tb,                         
showstringspaces=false
}}

\lstnewenvironment{python}[1][]
{
\pythonstyle
\lstset{#1}
}
{}

\newcommand\pythoninline[1]{{\pythonstyle\lstinline!#1!}}

\journal{Engineering Structures}

\newcommand{\sps}{\vspace{2ex}}
\pgfplotsset{compat=1.17}

\begin{document}

\begin{frontmatter}




\title{Self-learning locally-optimal hypertuning using maximum entropy, and comparison of machine learning approaches for estimating fatigue life in composite materials}


\renewcommand{\thefootnote}{\fnsymbol{footnote}}

\author{Miguel Díaz-Lago$^{a,}$\footnote[2]{M.Díaz-Lago and I. Ben-Yelun contributed equally to this work as first authors.}, Ismael Ben-Yelun$^{a,\dag}$, Luis Saucedo-Mora$^{a,b,c,}$\footnote[1]{Corresponding author\\ \hspace*{6mm}
\textit{Email address}: \texttt{luis.saucedo@upm.es} (Luis Saucedo-Mora).}, Miguel Ángel Sanz$^a$, Ricardo Callado$^a$, Francisco Javier Montáns$^{a,d}$}

\address{$^a$ E.T.S. de Ingeniería Aeronáutica y del Espacio, Universidad Politécnica de Madrid, Pza. Cardenal Cisneros 3, 28040, Madrid, Spain\\

\vspace{0.3cm}

$^b$ Department of Materials, University of Oxford, Parks Road, Oxford, OX1 3PJ, UK\\

\vspace{0.3cm}

$^c$ Department of Nuclear Science and Engineering, Massachusetts Institute of Technology,  MA02139, USA\\

\vspace{0.3cm}

$^d$ Department of Mechanical and Aerospace Engineering, Herbert Wertheim College of Engineering, University of Florida, FL32611, USA}

\begin{abstract}
Applications of Structural Health Monitoring (SHM) combined with Machine Learning (ML) techniques enhance real-time performance tracking and increase structural integrity awareness of civil, aerospace and automotive infrastructures. This SHM-ML synergy has gained popularity in the last years thanks to the anticipation of maintenance provided by arising ML algorithms and their ability of handling large quantities of data and considering their influence in the problem.

In this paper we develop a novel ML nearest-neighbors-alike algorithm based on the  principle of maximum entropy to predict fatigue damage (Palmgren-Miner index) in composite materials by processing the signals of Lamb Waves---a non-destructive SHM technique---with other meaningful features such as layup parameters and stiffness matrices calculated from the Classical Laminate Theory (CLT). The full data analysis cycle is applied to a dataset of delamination experiments in composites. The predictions achieve a good level of accuracy, similar to other ML algorithms, e.g. Neural Networks or Gradient-Boosted Trees, and computation times are of the same order of magnitude.

The key advantages of our proposal are: (1) The automatic determination of all the parameters involved in the prediction, so no hyperparameters have to be set beforehand, which saves time devoted to hypertuning the model and also represents an advantage for autonomous, self-supervised SHM. (2) No training is required, which, in an \textit{online learning} context where streams of data are fed continuously to the model, avoids repeated training---essential for reliable real-time, continuous monitoring.
\end{abstract}


\end{frontmatter}



\section{Introduction}

Civil infrastructures, built since ages, are supposed to be robust and safe to provide service to the daily life of millions of people during their operational lifespan. This concern is also present in more recent fields, such as the aerospace or the automotive sector, where the operation in highly aggressive dynamic environments and the necessary lightness and tight safety factors may increase the possibility of damage and a lack of safety during their service life. Therefore, it is fundamental to ensure the structural integrity of these systems, which must be maintained over time. Structural health monitoring (SHM) techniques are intended to undertake potential structural problems that may arise during the lifetime of these structures, either to perform a corrective maintenance e.g. replace a certain damaged part, or a predicted maintenance, trying to take advantage before the damage becomes relevant and thus providing added safety and reducing maintenance costs to the given infrastructure. A recent review of the many existing techniques in SHM has been done by Rocha et al. \cite{rocha2021sensors}, but in the last decades a large quantity of research has been performed \cite{doebling1998summary, sohn2003review, ko2005technology, farrar2007damage, song2009improved, ou2010structural, farrar2012structural}. This monitoring essentially consists in information---captured by sensors or similar devices---about the health of the component, that is, downstreaming data that have to be interpreted by the user.

The real-time data generation which is constantly monitored in SHM is a key feature in our digitalized world, where vast quantities of data are continuously shared, downloaded and processed at the same time---according to a research conducted by Holst, $79\cdot 10^{21}\,$bytes of data were generate worldwide in 2021 \cite{holst2021amount}. Furthermore, this existing data flow is taking place not only in terms of the available amount of data but also in terms of the velocity in which the information is generated, transferred, and demanded.

These two last claims have both enabled and likewise been enabled due to the use of data driven techniques such as Machine Learning to process these gathered data and make predictions of the  health state of the structure. This effective synergy has been sought in the development of this work, which addresses damage mechanics monitored in composite materials. Both topics introduce nonlinearities in terms of structural response, so the models used to simulate these effects become increasingly complex. Hence the necessity of coming up with a surrogate, physics-based ML model that bypasses the need of adjusting semi-analytical models for damage in composite materials, which has motivated the present work. These latest advances in ML modeling have allowed the use of SHM in other engineering areas like the one undertaken in this paper---a composite materials dataset from the NASA. There is already related research work dealing with this dataset which might be consulted \cite{saxena2011accelerated, chiachio2013energy, chiachio2013fatigue, peng2013novel}.

Regarding the physical-mathematical tools used to address this data-driven topic, the statistical focus and derivatives has been utilized as common approach---see Ko and Ni \cite{ko2005technology}---as well as autoreggresive (AR) models \cite{magalhaes2012vibration}. However, the latter method has the limitations of lacking external variables, which is a critical issue when trying to forecast in time series \cite{lin2020limitations}. In another attempt to apply predictive maintenance in (infra)structures, Gaussian Process (GP), probabilistic bayesian and transfer-bayesian models might be found in Wan and Ni \cite{wan2018bayesian}, Bull et al. \cite{bull2019probabilistic} and Ierimonti et al. \cite{ierimonti2021transfer}, with subsequent improvements in predictions due to the ability of the models to handle non-linear input data. Nevertheless, these models depend excessively on the train set i.e. the data used for their fitting, leading to non-unique solutions, and also present the main drawback of computation time of GP---of order $\mathcal{O}(n^3)$. Lastly, common ML techniques such as Neural Networks (NN) and Support Vector Machine (SVM) have been recently used for SHM applied to bridges \cite{finotti2019shm}, yet these algorithms require hypertuning, and more in particular, the input data of the model proposed in \cite{finotti2019shm} is purely autoreggresive, with the previously mentioned disadvantages that this kind of models pose. 

In this paper we propose a novel ML nearest-neighbors-alike algorithm which maximizes the entropy (i.e. amount of information) of the surrounding data points and makes a prediction through a convex combination of these selected neighbors---the set of points that maximizes the entropy. This algorithm is applied to a real dataset of delamination tests of composite layups to predict their fatigue life, and compared to other commonly used Machine Learning algorithms such as k-Nearest Neighbors, gradient boosting trees and Neural Networks. There are two key advantages with respect to the other algorithms: The first one is that all the neighbors parameters (i.e. the radii) are optimized \textit{during} the prediction, thus saving the need of finding a suitable hyperparameter for the model. The second one is the accomplishment of a real-time ML model, able to handle with new, continuously-generated data streams. The latter feature not only avoids repeated training of the model but also represents a supply of new data points that enriches the accuracy for further predictions---something which is essential for real-time and continuous monitoring.

This paper is organized as follows. Firstly, a theoretical review of composites and structural health monitoring (SHM) techniques is performed in Section~\ref{Sec:theoretical_framework}. Then, the maximum entropy algorithm is introduced in Section~\ref{Sec:max_entropy_algorithm}, explaining the steps of the algorithm itself and showing two examples as a proof of its functionality. Then, in Section~\ref{Sec:methodology} the proposed methodology to predict fatigue-life in composites is addressed. The full data analysis cycle is applied to the experiments conducted at Stanford Structures and Composites Laboratory (SACL) in collaboration with the Prognostic Center of Excellence (PCoE) of NASA Ames Research Center \cite{chiachio2013documentation}. Finally, the results are detailed in Section~\ref{Sec:results}, and several concluding remarks are outlined in Section~\ref{Sec:conclusions}.
\section{Theoretical Framework}
\label{Sec:theoretical_framework}

\subsection{Classical Laminate Theory}

Using some assumptions, the Classical Laminate Theory (CLT) \cite{jones2018mechanics} simplifies the complex nature of laminates in terms of mechanical properties of continua. Each ply has an orthotropic behaviour, from the notable difference in Young's moduli between the stiff fiber direction and its perpendicular plane, being the latter only determined by the resin properties. The CLT provides a method to calculate the continuum stiffness matrix of any laminate, which relates the forces and moments applied to a laminate plate to the in-plane strains $\varepsilon_{xx}, \varepsilon_{yy}, \gamma_{xy}$ and curvatures $\kappa_{xx}, \kappa_{yy}, \kappa_{xy}$. The stress tensor is obtained from the constitutive equations. The terms of the stiffness matrix will be later used as input features of the model, to feed it with elastic, i.e. physics-based, information.

The constitutive equations of each ply expressed in the main ply axes, $[Q]_{12}$, assuming stresses remain in-plane are 

\begin{equation}
	[\bm{\varepsilon}]_{12} = [Q]_{12} [\bm{\sigma}]_{12}
	\Rightarrow
    \begin{bmatrix}
    \varepsilon_1 \\
    \varepsilon_2 \\
    \gamma_{12}
    \end{bmatrix}
    =
    \begin{bmatrix}
    1/E_1 & -\nu_{12}/E_1  &  0\\
    -\nu_{12}/E_1 & 1/E_2 & 0\\
    0 & 0 & 1/G_{12}
    \end{bmatrix}_{12}
    \begin{bmatrix}
    \sigma_1 \\
    \sigma_2 \\
    \tau_{12}
    \end{bmatrix},
\end{equation}
where $E_1$, $E_2$, $\nu_{12}$, $G_{12}$ are the Young's  moduli, Poisson's ratio and shear modulus of the ply. Then, to use a common reference frame for all plies, we convert them to the laminate axes, $[Q]_{xy}$, through suitable rotation matrices. After that, applying the Kirchhoff hypotheses for plates and integrating across the $z-$coordinate of the laminate (with respect to the middle plane), the stiffness matrix of the laminate can be obtained: 



\begin{equation}
	\left[
	\begin{BMAT}[2pt,0pt,3cm]{c}{ccc;ccc}
		N_x \\ N_y \\ N_{xy} \\
		M_x \\ M_y \\ M_{xy}
	\end{BMAT}
	\right]
     = 
    \left[
    \begin{BMAT}[2pt,0pt,3cm]{ccc;ccc}{ccc;ccc}
        A_{11} & A_{12} & A_{16} & B_{11} & B_{12} & B_{16} \\
        A_{21} & A_{22} & A_{26} & B_{21} & B_{22} & B_{26} \\
        A_{61} & A_{62} & A_{66} & B_{61} & B_{61} & B_{66} \\
        B_{11} & B_{12} & B_{16} & D_{11} & D_{12} & D_{16} \\
        B_{21} & B_{22} & B_{26} & D_{21} & D_{22} & D_{26} \\
        B_{61} & B_{62} & B_{66} & D_{61} & D_{62} & D_{66} \\
    \end{BMAT}
    \right]
    \begin{bmatrix}
    \begin{BMAT}[2pt,0pt,3cm]{c}{ccc;ccc}
        \varepsilon_{xx} \\ \varepsilon_{yy} \\ \gamma_{xy} \\ \kappa_{xx} \\ \kappa_{yy} \\ \kappa_{xy}
    \end{BMAT}
    \end{bmatrix},
\end{equation}
where the terms of the three distinctive matrix boxes are

\begin{equation}
    A_{ij} = \sum_{k=1}^{n} Q_{ij}^k(z_k-z_{k-1}),
\quad
    B_{ij} = \frac{1}{2}\sum_{k=1}^{n} Q_{ij}^k(z_k^2-z_{k-1}^2),
\quad
    D_{ij} = \frac{1}{3}\sum_{k=1}^{n} Q_{ij}^k(z_k^3-z_{k-1}^3),
\end{equation}
being $[Q]^k$ the constitutive matrices of each ply $k$ in laminate axes, and $z_k$ are the locations of the plies relatives to the middle plane. This is the stiffness matrix of the laminate. Submatrix A is called \emph{extensional stiffness matrix} and relates extensional and shear forces and strains. Submatrix B is the \emph{coupling stiffness matrix}, since it relates bending strains with extensional and shear forces and vice-versa. Lastly, submatrix D is called the \emph{bending stiffness matrix}, for it relates the curvature with the bending moments. For \emph{symmetric} laminates, there is an extension-bending decoupling, i.e. $B_{ij} = 0$. For \emph{balanced} laminates extension and shear forces and strains are decoupled, $A_{16} \approx A_{26} \approx 0$.

\subsection{Structural Health Monitoring and Lamb Waves}

\emph{Structural Health Monitoring} (SHM) is defined as the process of implementing a damage detection strategy for aerospace, civil and mechanical engineering  infrastructures \cite{farrar2007introduction}. Damage in laminate components like delaminations, cracks, or porosity, can be produced during manufacturing or during service life, and grow damaging the mechanical properties of the component, leading to ultimate failure. Defects may also appear due to accidents like impacts, and may even not be visible in the surface of the component. Therefore, certification and maintenance tasks need reliable non-destructive testing (NDT) methods to check the health of laminate components.

Ultrasonic testing is the most widely spread NDT method for their accuracy and simplicity. Mechanical waves are introduced in the laminate via a transducer and propagate through the thickness of the laminate. When encountered with defects (heterogeneity in the material) they attenuate or get refracted, causing a different received signal than if the material had not had any defects. Other techniques like Phased Arrays \cite{mcnab1987ultrasonic}, Termography \cite{sakagami2002applications} and Optical NDT techniques \cite{zhu2011review} are also widely used.

As opposed to traditional ultrasonic testing, where ultrasounds travel perpendicularly to the laminate plane, Lamb waves travel within the laminate plane. This makes the testing less tedious and time consuming as larger areas can be examined at once. It has been proven that Lamb Waves scatter in defects like cracks or delaminations of the material, thus changing their propagation pattern and increasing dissipation through friction at those effects \cite{guo1993interaction, worlton1956ultrasonic}. With that, properties of the received wave such as amplitude and phase differ from the emitted one.

By comparing properties of the received Lamb wave signals of a given measurement to that of the equivalent measurement with no damage, it could be possible to obtain meaningful information about the defects present in the material. This principle will be used in the construction of the model to extract features containing damage information.
\section{Maximum Entropy Algorithm}
\label{Sec:max_entropy_algorithm}

\subsection{Mathematical Formulation}
The proposed Maximum Entropy algorithm is an algorithm in the family of k-Nearest Neighbors. The number of neighbors, $k$, is chosen by the algorithm for every point to predict by maximizing the mean entropy of the set. Contrary to other k-Nearest Neighbors, this parameter is different for every point to predict and it is not defined by the user, but obtained by the algorithm. Then, the algorithm calculates the prediction by interpolating the coordinates in the input space through convex combination.

\subsubsection{Convex Subset Definition}
\begin{figure}[H]
    \centering
    \includegraphics[width = .8\textwidth]{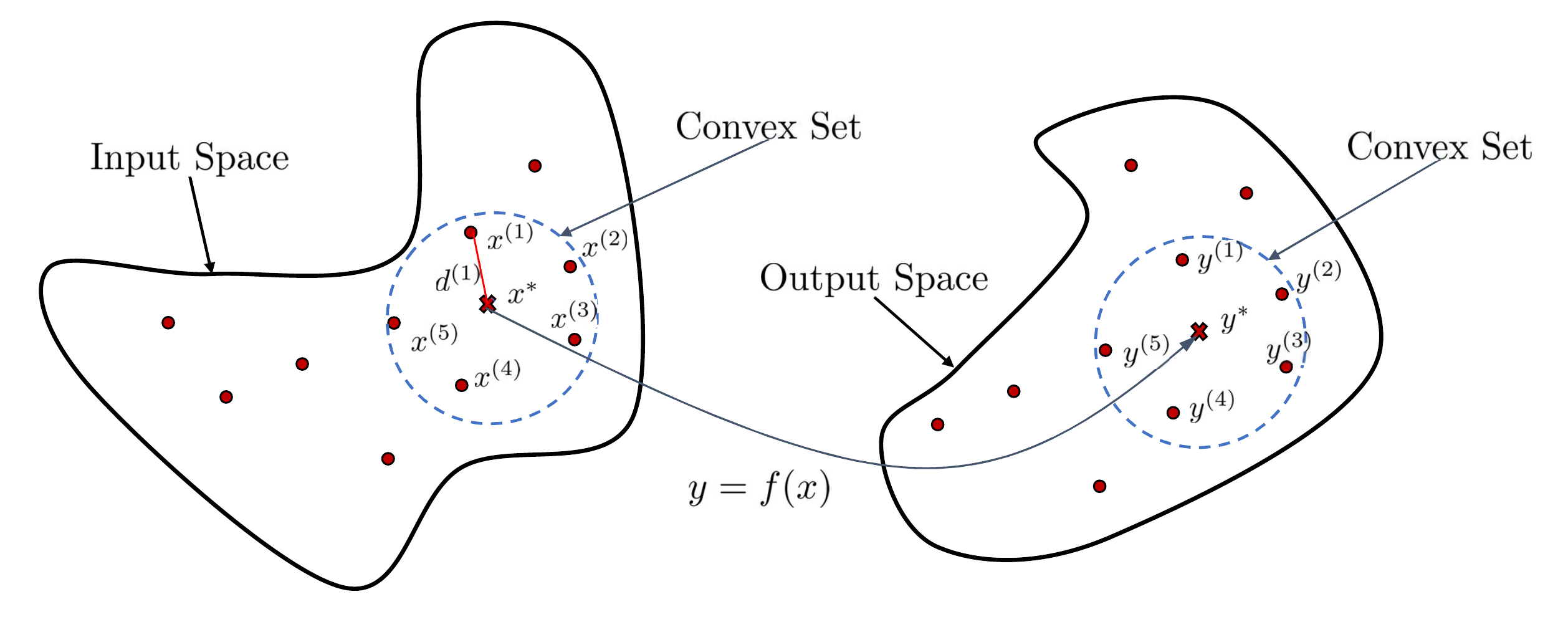}
    \caption{Illustration of the convex spaces defined within the input and output spaces.}
    \label{convex}
\end{figure}

The input space or feature space can be considered an space $\mathbb{R}^n$ where $n$ is the number of features of the data. The first step of the algorithm is to define a convex subset of the data around the point to predict. The size of this subset is not equal for all the points and will be self-determined by the maximum entropy principle. Another derived convex subset will also be defined in the output or image space, containing the images of the same data points previously added to the convex set in the original space. Figure \ref{convex} illustrates the concept.

\figurename~\ref{convex} represents both input and output spaces and their respective convex subsets. They are defined around the point to predict, and will contain a number of neighbors previously unknown.

Whether a point is considered part of the convex set or not, will be determined by the \emph{Radial Basis Function} (RBF) with respect to the point to predict defined in Equation (\ref{rbf}) below. If the RBF value is higher than a certain threshold, this point will be included in the subset. For the point which label is to be predicted, $\mathbf{X}^*$, the RBF of every point $\mathbf{X}^{(i)}$ in the set is interpreted as a probability distribution and defined as

\begin{equation}\label{rbf}
    \mathrm{RBF}^{(i)}(\mathbf{X}^*;h) = \exp{\left( -\frac{||\mathbf{X}^{(i)}-\mathbf{X}^*||^2}{h^2} \right)},
\end{equation}
where $\mathbf{X}^*$ and $\mathbf{X}^{(i)}$ are vectors containing the whole set of features: $\mathbf{X}^* = [x^*_1, x^*_2 ... x^*_n]$ and $\mathbf{X}^{(i)} = [x_1^{(i)}, x_2^{(i)} ... x_n^{(i)}]$. Therefore $||\mathbf{X}^{(i)}-\mathbf{X}^*||$ represents their distance in $\mathbb{R}^n$. The value $\mathrm{RBF}^{(i)}$ quantifies the distance from the data point to the point to predict, damped or amplified by $h$. This parameter will act as the ``radius'' of the convex set. By increasing it of decreasing it more or less data point will pass the threshold and be counted in the convex set. Therefore, for the threshold $t$, the convex subset $C$ is defined as:
\begin{equation}
    C = \lbrace \mathbf{X}^{(i)}:\mathrm{RBF}^{(i)}(h)> t \rbrace.
\end{equation}

\subsubsection{Maximization of Mean Entropy}
 Shannon's Principle of Maximum Entropy \cite{shannon1948mathematical} will be used to choose the local optimal value $h^*$ of $h$ and therefore, the size of the convex set. Once an initial convex subspace has been defined, the maximization of mean entropy is achieved by maximizing the sum of Gibbs-type entropy contributions of all points in the convex subset $C$:

\begin{equation}\label{meanentropy}
    h^* = \underset{h}{\arg \max} \left( - \frac{1}{m_C} \sum_{i \in C} \mathrm{RBF}^{(i)}(h) \log \left( \mathrm{RBF}^{(i)}(h) \right) \right),
\end{equation}
where $m_C$ is the amount of data points in the convex subset excluding $\mathbf{X}^*$. After finding this value $h^*$, the RBFs of the points are once again computed and filtered through a threshold. Then, the final convex subset is defined. 

\subsubsection{Convex Combination}
Once the convex subset is defined, the interpolation weights are calculated. This is done through the convex combination \cite{tyrrell1970convex}. Given a set of points within a convex space $C$, $\lbrace \mathbf{X}^{(i)} \rbrace$ the coordinates or features of the point which label is to be predicted $\mathbf{X}^*$ are estimated as:
\begin{equation}\label{xhat}
    \hat{ \mathbf{X}}  = \sum_{i \in C}u^{(i)} \cdot\mathbf{X}^{(i)},
\end{equation}
where $\hat{\mathbf{X}}$ is the convex combination approximation, subject to the constraints: 

\begin{equation}\label{c1}
    u^{(i)} \geq 0, \; \forall i,
\quad
    \sum_{i} u^{(i)} = 1.
\end{equation}

If $\mathbf{X}^* = \hat{\mathbf{X}}$ we expect the most accurate prediction, otherwise an error should be expected. Notice that the same interpolation weights $\mathbf{u}$ are used to interpolate every feature for a given point to predict $\mathbf{X}^*$. Conditions (\ref{xhat}) and (\ref{c1}) above can be expressed as matrix set of equations plus the non-negative constraint:
\begin{equation}\label{system}
    \begin{bmatrix}
        x^{(1)}_1 & x^{(2)}_1 & x^{(3)}_1 & \dots & x^{(k)}_1 \\
        x^{(1)}_2 & x^{(2)}_2 & x^{(3)}_2 & \dots & x^{(k)}_2 \\
        \vdots & \vdots & \vdots & \ddots & \vdots\\
        x^{(1)}_n & x^{(2)}_n & x^{(3)}_n & \dots & x^{(k)}_n\\
        1 & 1 & 1 & \dots & 1
    \end{bmatrix}
    \begin{bmatrix}
        u^{(1)} \\ u^{(2)}  \\ \vdots \\ u^{(k)}
    \end{bmatrix}
    =
    \begin{bmatrix}
        x^*_1 \\x^*_2 \\ \vdots \\ x^*_n \\ 1
    \end{bmatrix},\; 
    u^{(i)} \geq 0,
\end{equation}
where $k$ is the number of data points in the convex set, which may be different for every point to predict. The weights $\mathbf{u}$ are calculated and forced to be positive. In general, the dimensions of the matrix, $n+1$ and $k$ will be different and the system will not be square. For the usual case of $n+1>k$ the system will be over-constrained and there will not be any exact solution. Also, as the number of features, $n$, gets bigger, the constraint in Equation (\ref{c1}) will be harder to satisfy.

\subsubsection{Calculation of the Interpolation Weights}
Generally, the non-square system in Equation (\ref{system}) will be expressed as 
\begin{equation}
    \mathbf{K}\mathbf{u}=\mathbf{b},\; u^{(i)} \geq 0.
\end{equation}

Any numerical (e.g. iterative) method could be used for its solution such as the Moore-Penrose pseudoinverse. Herein, one iteration in the solution of the system is:

\begin{equation}
    \mathbf{u}^{\prime} = f(\mathbf{K},\mathbf{u},\mathbf{b}).
\end{equation}

For every iteration the convex approximation of $\mathbf{X}^*$, $\hat{\mathbf{X}}$, will be calculated as in Equation (\ref{xhat}). The iterative calculation of the weights will continue until the error, which we define below, is lower than a given tolerance.

\subsubsection{Calculation of the Error}
The error in the target is estimated to be similar to the error in the interpolation of the data which label is to be predicted, and it is defined as the norm of difference between the real point to predict and its interpolation, normalized by the norm of the interpolation:

\begin{equation}\label{error}
    \mathrm{error} = \frac{||\mathbf{X}^* - \hat{\mathbf{X}}||}{||\mathbf{X}^*||}, \; ||\mathbf{X}^*|| \neq 0
\end{equation}

A series of stopping conditions involving the amount of iterations and the value of the error will evaluate whether to continue iterating over $\mathbf{u}$ or to stop.

\subsubsection{Calculation of the Prediction}
Once convergence has been reached and the interpolation weights are obtained, the predictions of the labels are calculated. For regression, the predictions are (scalar case)

\begin{equation}\label{predreg}
    \hat Y = \sum_{i \in C} u^{(i)} Y^{(i)},
\end{equation}
where $\hat Y$ is the prediction for $\mathbf{X}^*$ and $Y^{(i)}$ are the labels of the data points in the convex subset $C$. For classification, the class label of the point to predict is just the statistical mode of the classes of the neighbors in the convex set:

\begin{equation}\label{predclas}
    \hat{Y} = \mathrm{mode} \left(Y^{(i)}\right),\; i \in C.
\end{equation}

In this case, the interpolation weights $u$ are not used, thus saving a step in terms of computation time. This formulation allows to calculate a single point to predict, $\mathbf{X}^*$. For multiple outputs $\hat{\mathbf{Y}} = \left\lbrace\hat{Y}_1, \hat{Y}_2, ... \hat{Y}_t\right\rbrace$, Equations (\ref{predreg}) or (\ref{predclas}) should be applied $t$ times. A more detailed description of the algorithmic and the workflow diagram is included in \ref{Apendx:implementation}.

\subsection{Toy Examples}

Two tests, one for regression and one for classification are used to test the approach. They are good practical examples to demonstrate the capabilities of the code.

\subsubsection{Regression Toy Example}
The aim of this toy example is to predict a given function $Y: \mathbb{R}^2 \to \mathbb{R}^2 $. The features $x_1$ and $x_2$ are assured to vary in the range $[0,1]\times[0,1]$. 500 data points are randomly scattered along the input space. The ground-truth functions to be predicted are:

\begin{equation}
    y_1 = \cos{\left(\dfrac{x_1}{0.3}\right)}\sin{\left(x_2\right)},
    \qquad
    y_2 = x_1x_2
\end{equation}

\begin{figure}[H]
    \centering
    \includegraphics[width = 0.85\textwidth, trim={0cm 0cm 0cm 0.9cm}, clip]{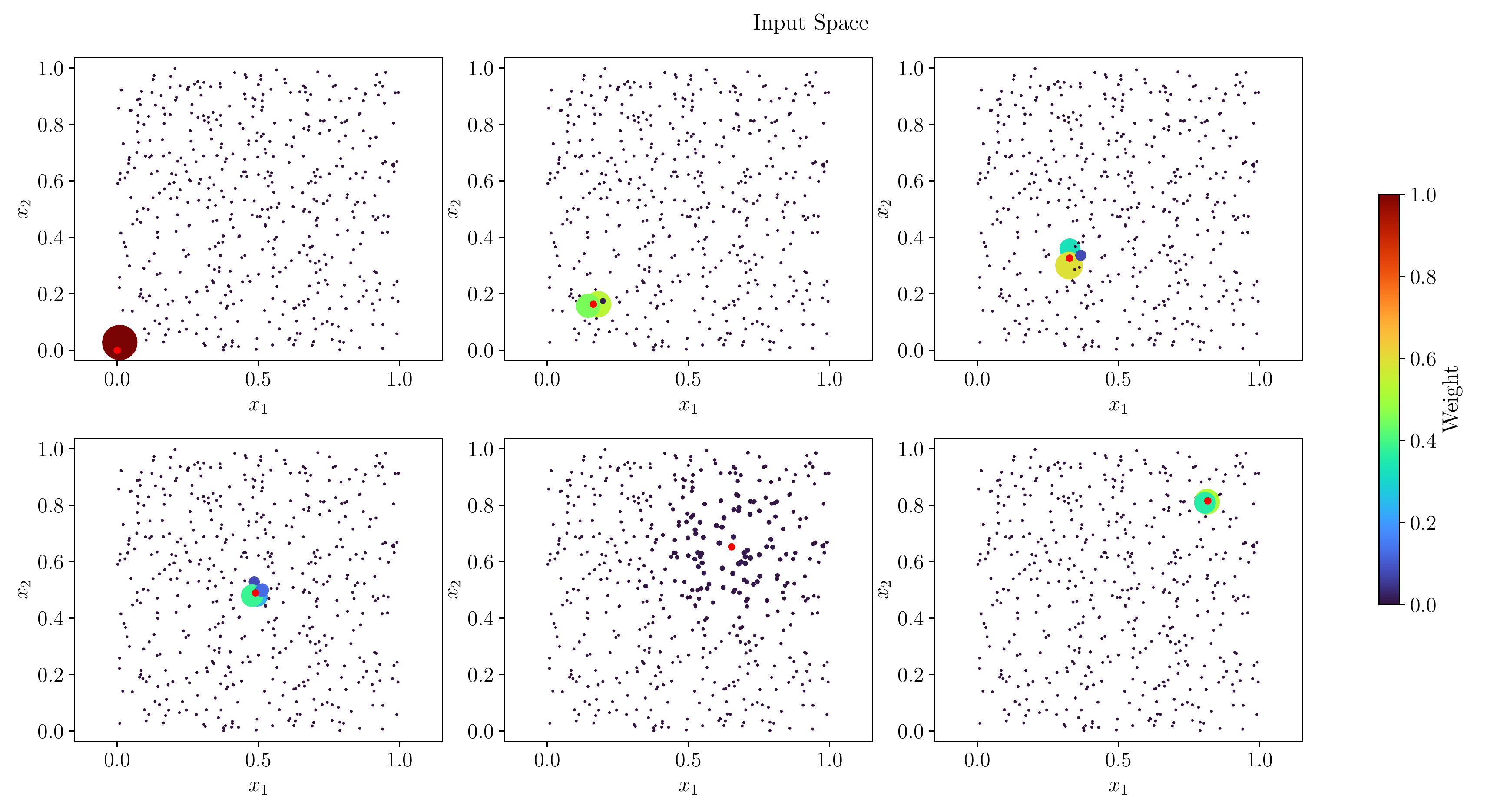}
    \caption{Input space of the regressor toy example.}
    \label{testreg_input}
\end{figure}

The evaluation points will be 50 points equally spaced along the line $x_1 = x_2$. Figure \ref{testreg_input} displays the input space for six different points across the diagonal for which the labels are to be predicted. The specific point is marked in red in its corresponding figure frame, whilst the rest of dots represent the data points that have been provided to the algorithm along with their labels, obviously the same in all six frames. Their color and area have been scaled to display the weight each of them carries in the interpolation of the label.

\begin{figure}[htb]
    \centering
    \includegraphics[width = .9\textwidth]{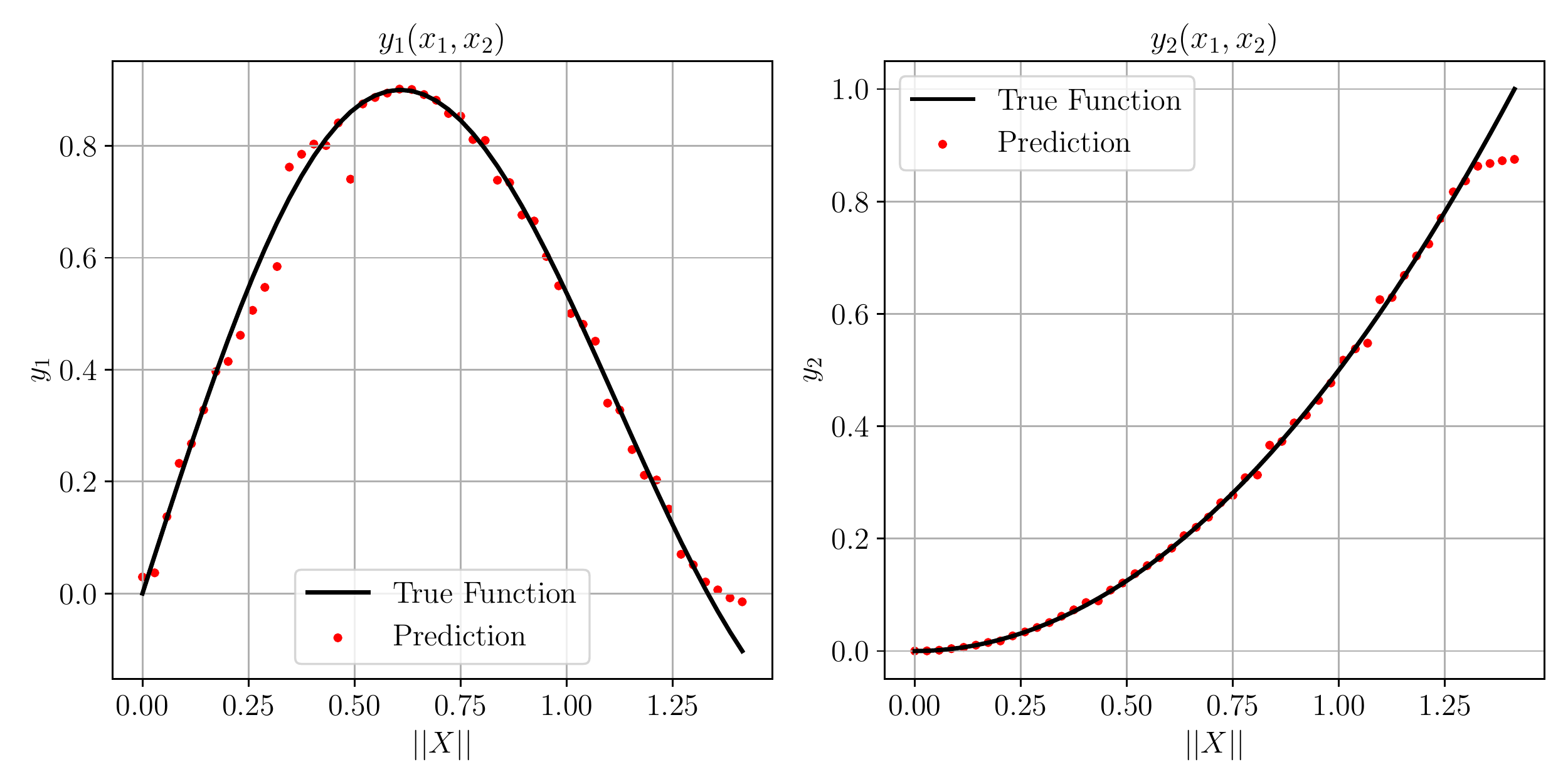}
    \caption{Output space of the regressor toy example. Function $y_1$ in the left, $y_2$ on the right.}
    \label{testreg_output}
\end{figure}

As can be seen, depending on the point to predict the number of neighbors varies: from very few with high weights like in the top left case, to many neighbors with low weights like in the central bottom case. The output is depicted in Figure \ref{testreg_output}.

In Figure \ref{testreg_output}, both output values have been depicted against the norm of the features, sticking to the true function with a certain margin of error. It can be confirmed that the areas where the error is greater correspond to those where the points to predict have further neighbors or to those near the boundaries. This often happens near the boundaries of the input space and it is an effect of the boundary conditions \cite{fernandez2004imposing}, which increases the error.

\subsubsection{Classification Toy Example}

\begin{figure}[htb]
    \centering
    \includegraphics[width = 0.9\textwidth]{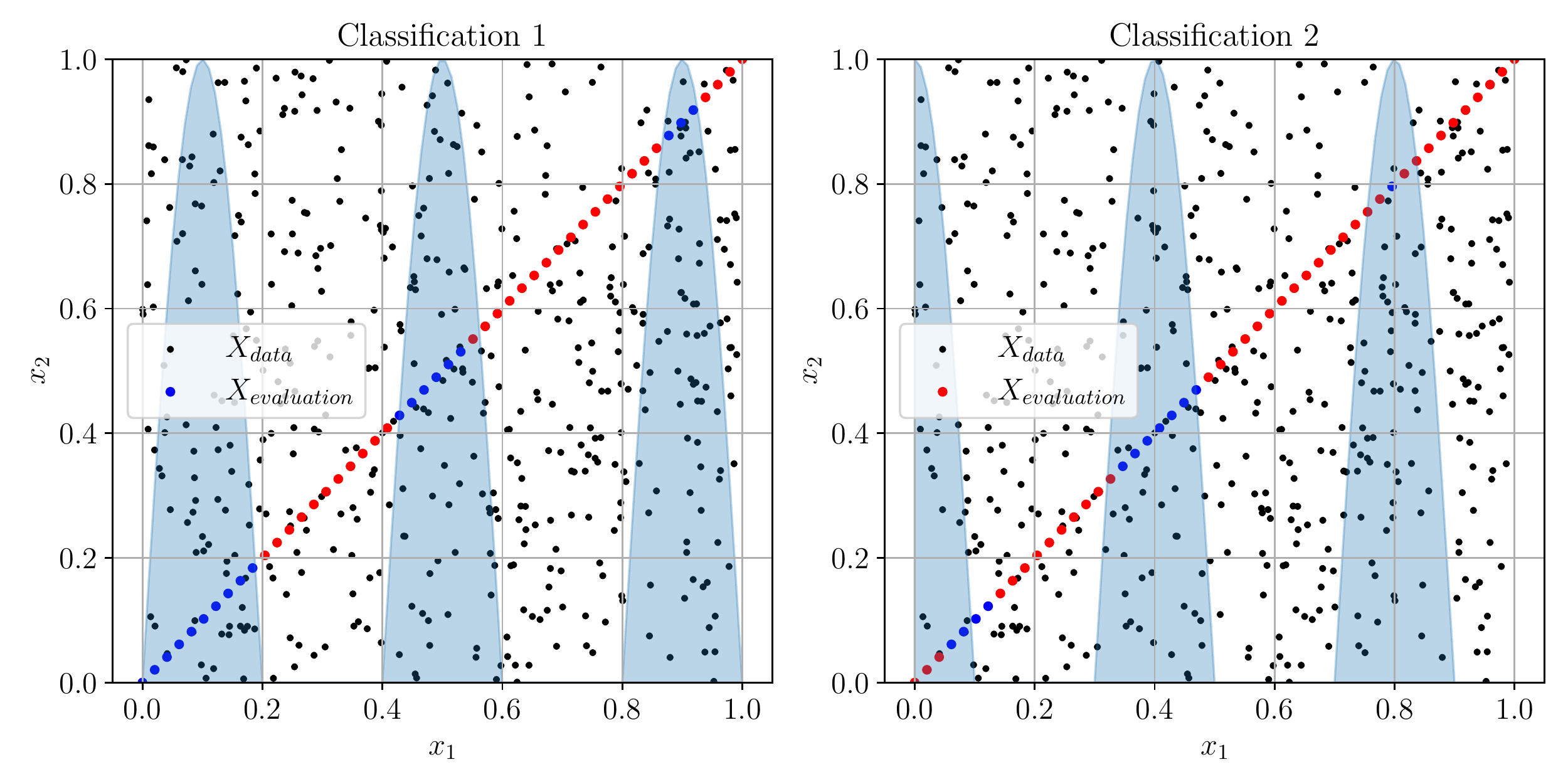}
    \caption{Classifier test example. Classification $y_1$ in the left, $y_2$ on the right.}
    \label{testclas}
\end{figure}

Now, two binary classifications have been performed. The interval from $\mathbb{R}^2$, $[0,1]\times[0,1]$ is likewise chosen as input space. The space has been labeled as 0 or 1 as in:

\begin{equation}
    y_1= \left\{ \begin{array}{ll}
             1 &   \mathrm{if}  \quad x_2 \geq \sin{\left( 5\pi x_1 \right)} \\
             \\ 0 &   \mathrm{otherwise}\\
             \end{array}
   \right.
\qquad
    y_2= \left\{ \begin{array}{ll}
             1 &   \mathrm{if}  \quad x_2 \geq \cos{\left( 5\pi x_1 \right)} \\
             \\ 0 &   \mathrm{otherwise} \\
             \end{array}
   \right.
\end{equation}
Then, 500 data points with their corresponding label have been scattered along the space, labelled accordingly, and passed to the algorithm. The task for the algorithm is to classify 50 points along the diagonal of the space. The results are presented in Figure \ref{testclas}.

In Figure \ref{testclas} colors are just placeholders for the labels: the pale blue area and the blue dots correspond to label 0, true value and predicted value respectively. The white area and red predictions correspond to label 1, also to true value and predicted value respectively. Like before, the black dots represent the data points. The diagonal line of red and blue dots are the points which label is to be predicted. As can be seen, almost all points that have fallen into the blue area ($\mathrm{label}=0$) are labeled as 0 (blue) and all the points in the white area ($\mathrm{label}=1$) as 1 (red), up to a degree of accuracy.
\section{Methodology}
\label{Sec:methodology}

As a practical real-world application, the Maximum Entropy algorithm will be used to predict the fatigue damage index from a series of Lamb wave measurements over CFRP coupons. A Machine Learning study is performed and we also compare the accuracy of the algorithm with other state-of-the-art Machine Learning models: Fully Connected (FC) Neural Networks from \texttt{keras} \cite{chollet2015keras}, LightGBM \cite{ke2017lightgbm}, XGBoost \cite{Chen:2016:XST:2939672.2939785} and Weighted k-Nearest Neighbors \cite{dudani1976distance} from \texttt{scikit-learn} \cite{scikit-learn}.

\subsection{Dataset}

\begin{figure}[H]
    \centering
    \includegraphics[width=.95\textwidth]{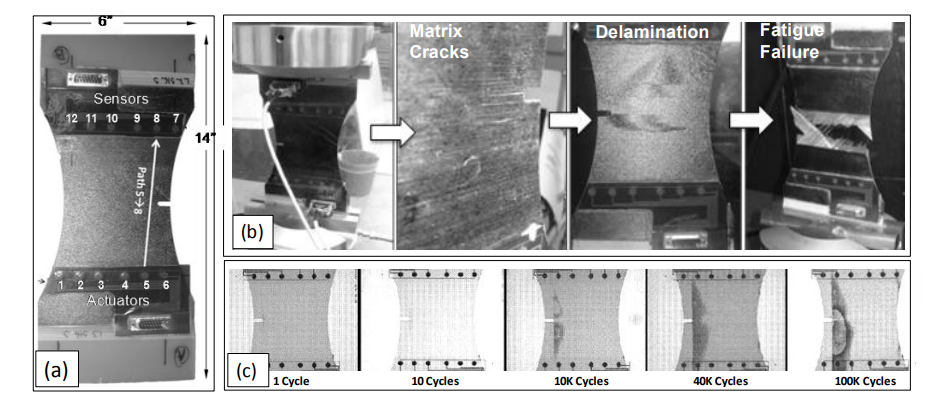}
    \caption{Experimental setup. Figure extracted from \cite{saxena2011accelerated}.}
    \label{exp}
\end{figure}

The experiments consisted in run-to-failure fatigue cycles over CFRP coupons. The load applied was a 80\% of the ultimate load, at frequency $f = 5\,\mathrm{Hz}$ and stress ratio of $\sigma_{\max}/\sigma_{\min} = 0.14$---where $\sigma_{\max}$ and $\sigma_{\min}$ are the maximum and minimum Cauchy stresses of the (fatigue) excitation load, respectively. At regular intervals in the number of cycles, the coupon was disassembled from the testing machine and the measurement was made. Tests were conducted at Stanford Structures and Composites Laboratory (SACL) in collaboration with the Prognostic Center of Excellence (PCoE) of NASA Ames Research Center. The results are public and available at the NASA Prognostics Data Repository webpage \cite{chiachio2013documentation}. In Figure \ref{exp}  the experimental setup is depicted.

The coupons are Torayca T700G uni-directional carbon-prepregs. Three different symmetric layups were used: $Layup 1: [90_2/45/-45]_{2S}$, $Layup 2: [0/90_2/45/\hbox{-45}/90]_S$ and $Layup 3: [90_2/45/-45]_{2S}$. The coupons were 15.24 cm x 25.4 cm with a dogbone geometry and a notch to induce stresses so the fatigue damage would originate there \cite{saxena2011accelerated}. The properties of the plies are given in \tablename~\ref{ply}.

\begin{table}[H]
    \centering
    {\small
    \begin{tabularx}{\textwidth}{|X|X|X|X|X|X|X|X|} 
        \hline
        $E_x$ [GPa] & $E_y$ [GPa] & $\nu_{xy}$ [-] & $\nu_{yz}$ [-] & $G_{xy}$ [GPa] & $G_{yz}$ [GPa] & $t$ [mm]\\
        \hline
        137.5 & 8.4 & 0.309 & 0.5 & 6.2 & 3.092 & 0.132 \\ 
        \hline
    \end{tabularx}}
    \caption{Mechanical properties of the ply \cite{chiachio2013energy}.}
    \label{ply}
\end{table}

The coupons had embedded two six-PZT-sensor SMART Layer® from Acellent Technologies, Inc. Six sensors acted as emitters and the other six as receivers---see \figurename~\ref{exp}. They sent Lamb wave signals that traveled across the coupon and were received at the other end. Each emitter was able to send signals in seven different frequencies. There were $6\times 6 = 36$ paths (every combination of emitter and receiver) and a total of $36\times 7 = 252$ channels (each path at 7 different frequencies). In the available data, each entry contained the Lamb Waves signals emitted and received in each channel as well as other information like the number of cycles at which it had been taken, the \emph{condition} and other information. The \emph{condition} determines whether the measurement was taken with the coupon disassembled from the traction machine (\emph{traction free}), still assembled but with no load applied, (\emph{clamped}) or with a constant load applied, (\emph{loaded}). Lastly, to undertake this dataset from the data analytics point of view, we parsed and formatted the (raw) data to provide them a meaningful form. The next steps are unique and within the novelty of the proposed algorithm.

\subsubsection{Palgrem-Miner's Damage Index}

Several articles have already used the dataset to predict fatigue damage \cite{saxena2011accelerated, chiachio2013energy, chiachio2013fatigue,  peng2013novel}. In them, some features of the signal with respect to the baseline---the measurement at 0 cycles---were extracted and used to predict different damage properties. A similar strategy is followed herein to predict the Palgrem-Miner's damage index \cite{miner1945cumulative} of the coupons

\begin{equation}
    D = \sum_{i} \frac{n_i}{N_i},
\end{equation}
where $n_i$ represents the number of cycles suffered at a certain frequency $i$ and $N_i$ the number of cycles at frequency $i$ at which the material is known to fail. The experiments are run until failure, hence $N_i$ is known for each coupon. Since the fatigue test is run with a unique frequency, i.e. $f=5\,$Hz, the previous expression is reduced to a single term $D = n/N$. Therefore, this index varies from $D =0$, when the material is free of defects, to $D=1$, when material is expected to fail. Then, the index provides information of its health within these values. The fatigue damage index is computed for every measurement and it is the the variable sought to predict by the model.

\subsection{Data Analysis Workflow}

The steps in the Machine Learning workflow are covered to study the data and extract useful features of the Lamb Wave signals to be fed into the models. Table \ref{amount_data} shows the number of coupons for each layup and the number of measurements for each coupon.

\begin{table}[H]
    \centering
    {\small
    \begin{tabularx}{\textwidth}{|l|X|X|X|X|X|X|X|X|X|X|X|X|} 
        \hline
        \multicolumn{13}{|c|}{Number of measurements}\\
        \hline
        \multicolumn{1}{|c|}{Layup} & \multicolumn{4}{|c|}{L1} & \multicolumn{4}{|c|}{L2} & \multicolumn{4}{|c|}{L3}\\
        \hline
        Coupon & S11 & S12 & S18 & S19 & S11 & S17 & S18 & S20 & S11 & S13 & S18 & S20 \\
        \hline
         & 147 & 109 & 181 & 83 & 129 & 109 & 109 & 108 & 122 & 147 & 136 & 124 \\
        \hhline{~------------}
        \multirow{1}{*}{PZT} & \multicolumn{4}{|c|}{516}& \multicolumn{4}{|c|}{452}& \multicolumn{4}{|c|}{526}\\
        \hhline{~------------}
        & \multicolumn{12}{|c|}{1492}\\
        \hline
    \end{tabularx}}
    \caption{Amount of PZT measurements in each coupon, layup and in total. }
    \label{amount_data}
\end{table}

\begin{figure}[H]
    \centering
    \includegraphics[width =\textwidth]{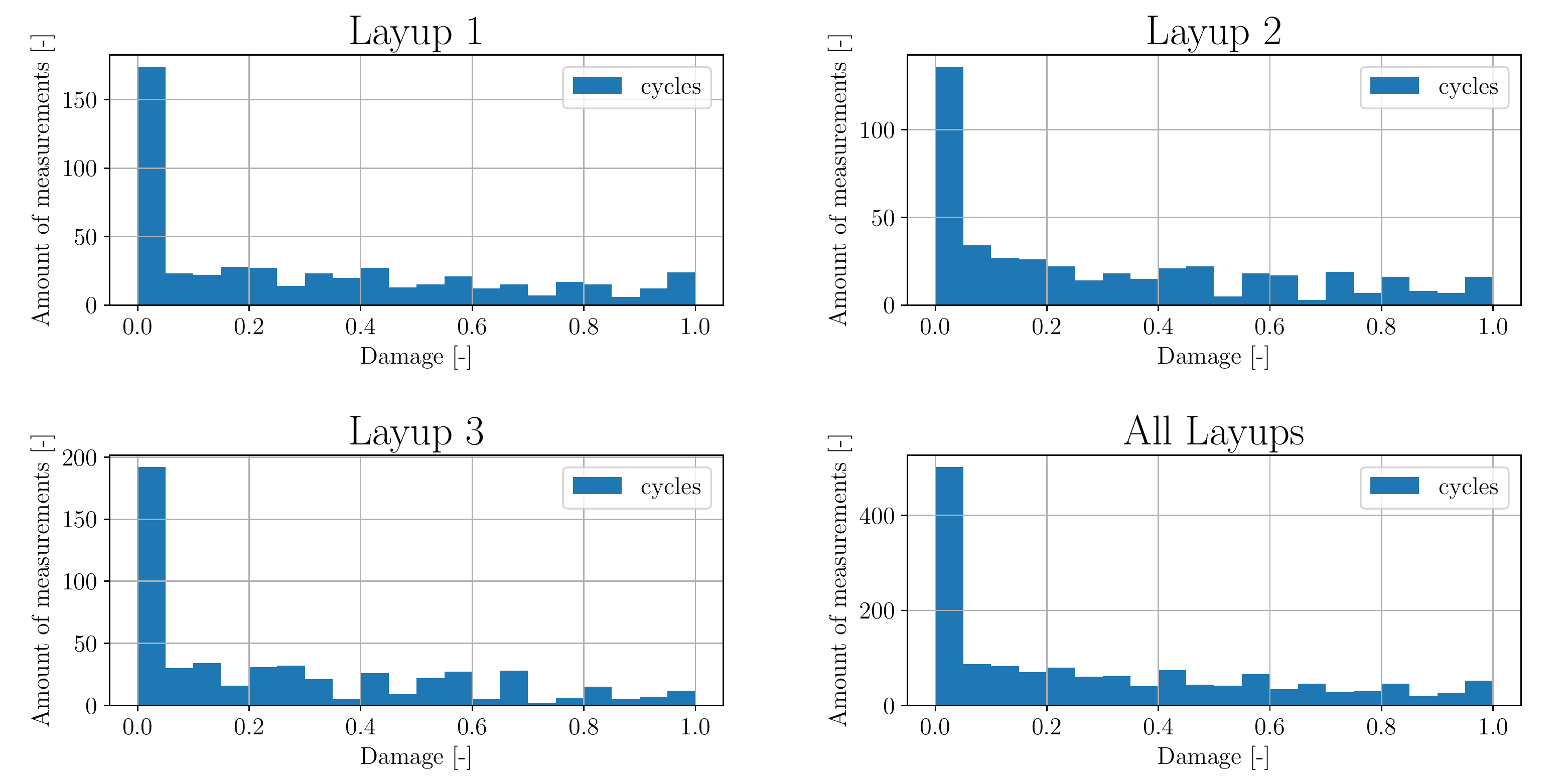}
    \caption{Amount of measurements against damage index for each layup and all layups.}
    \label{hist}
\end{figure}

As it can be seen, for modern day ML standards, the dataset can be considered small. In addition to this, Figure \ref{hist} shows the distribution of measurements against the fatigue damage index. It is observed that it is shifted---skewed---towards the low range of the fatigue damage index.  This makes it harder for the models to accurately predict examples in the high end of the range.

\begin{figure}[H]
    \centering
    \includegraphics[width = \textwidth]{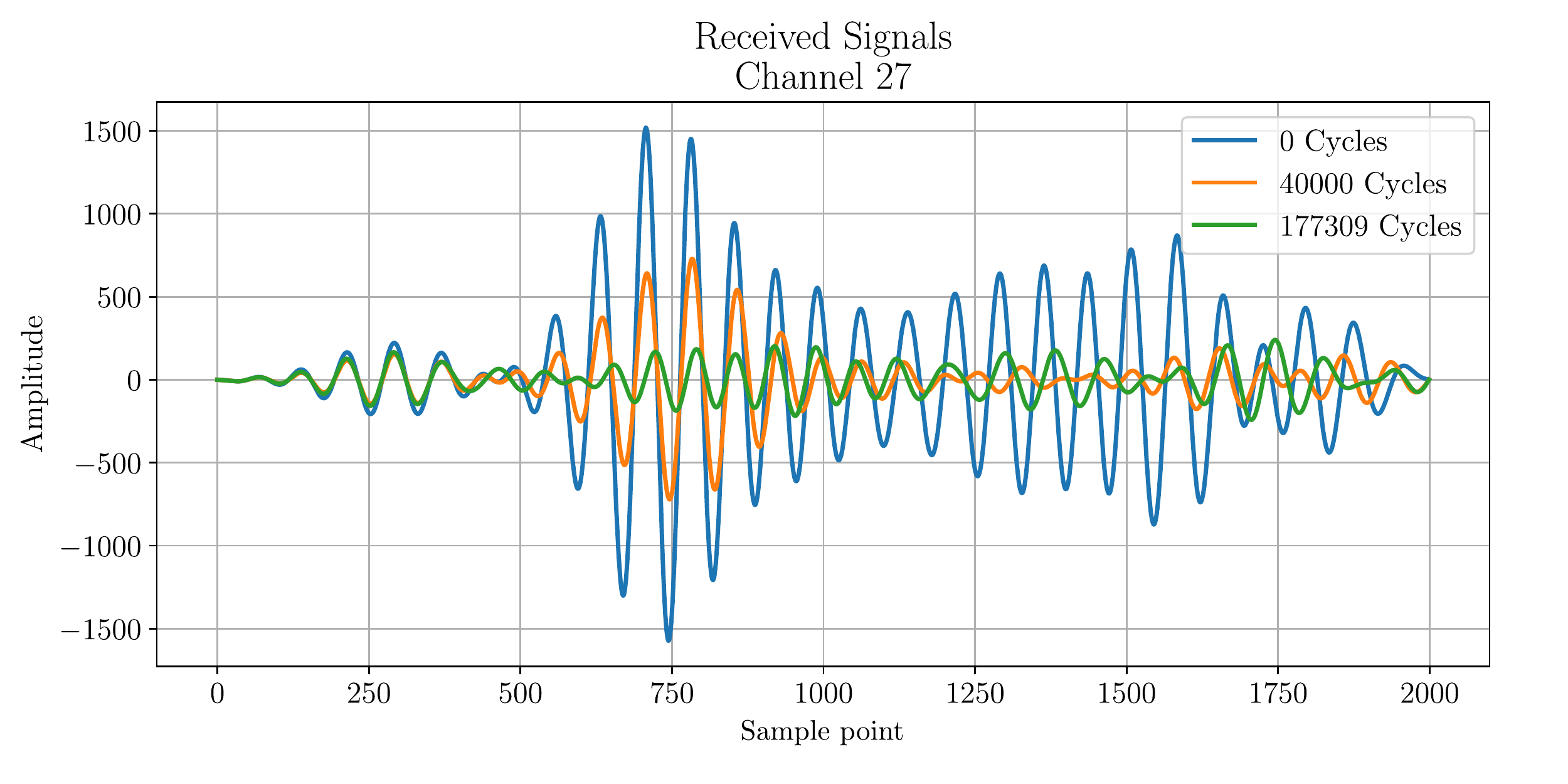}
    \caption{Received signals in coupon L1S11 and channel 27 at 0, 40000 and 177309 cycles.}
    \label{sig1}
\end{figure}

Regarding the available data from the Lamb wave signals, it is observed that, for most of the channels, a decrease in amplitude is noticeable as the number of cycles at which the measurement was taken increases. Figure \ref{sig1} depicts received signals for the same channel at different cycles.

In this figure it can observed that there is an amplitude decay and a subtle phase change (at the end of the wave) as the number of cycles increases. From this observation it could be inferred that the Lamb waves get scattered and lose energy (hence the decrease in amplitude) as they encounter defects such as delaminations and cracks within the material. The higher the number of suffered cycles, the more defects will be present and the amplitude of the signals will be smaller. In order to quantify this, two features are extracted: the power ratio and correlation coefficient with respect to the baseline. For the channel $c_i$, the power ratio is computed as
\begin{equation}
    pw\_ c_i = \frac{P_i}{P_{i}^{\mathrm{baseline}}},
\end{equation}
where the power for measurement $i$, $P_i$ is computed using the formula of the power of a non-periodic discrete signal \cite{chaparro2018signals}

\begin{equation}\label{power}
    P_i =  \frac{1}{2N+1} \sum_{n=-N}^N x_i^2(n), 
\end{equation}
being $x_i$ the signal of the Lamb wave sampled in a total of $2N+1$ points. Likewise, the Pearson's correlation coefficient between a measurement and its respective baseline for channel $c_i$ is: 

\begin{equation}
    cc\_c_i = \frac{\mathrm{Cov}(x_i,x_i^{\mathrm{baseline}})}{\sqrt{\mathrm{Var}(x_i)\mathrm{Var}(x_i^{\mathrm{baseline}})}}.
\end{equation}

These features are calculated for all measurements in the dataset. In Figure \ref{p27} the power ratio of an example channel for the data of one coupon is displayed, joint and separated by conditions:

\begin{figure}[H]
    \centering
    \includegraphics[width =\textwidth]{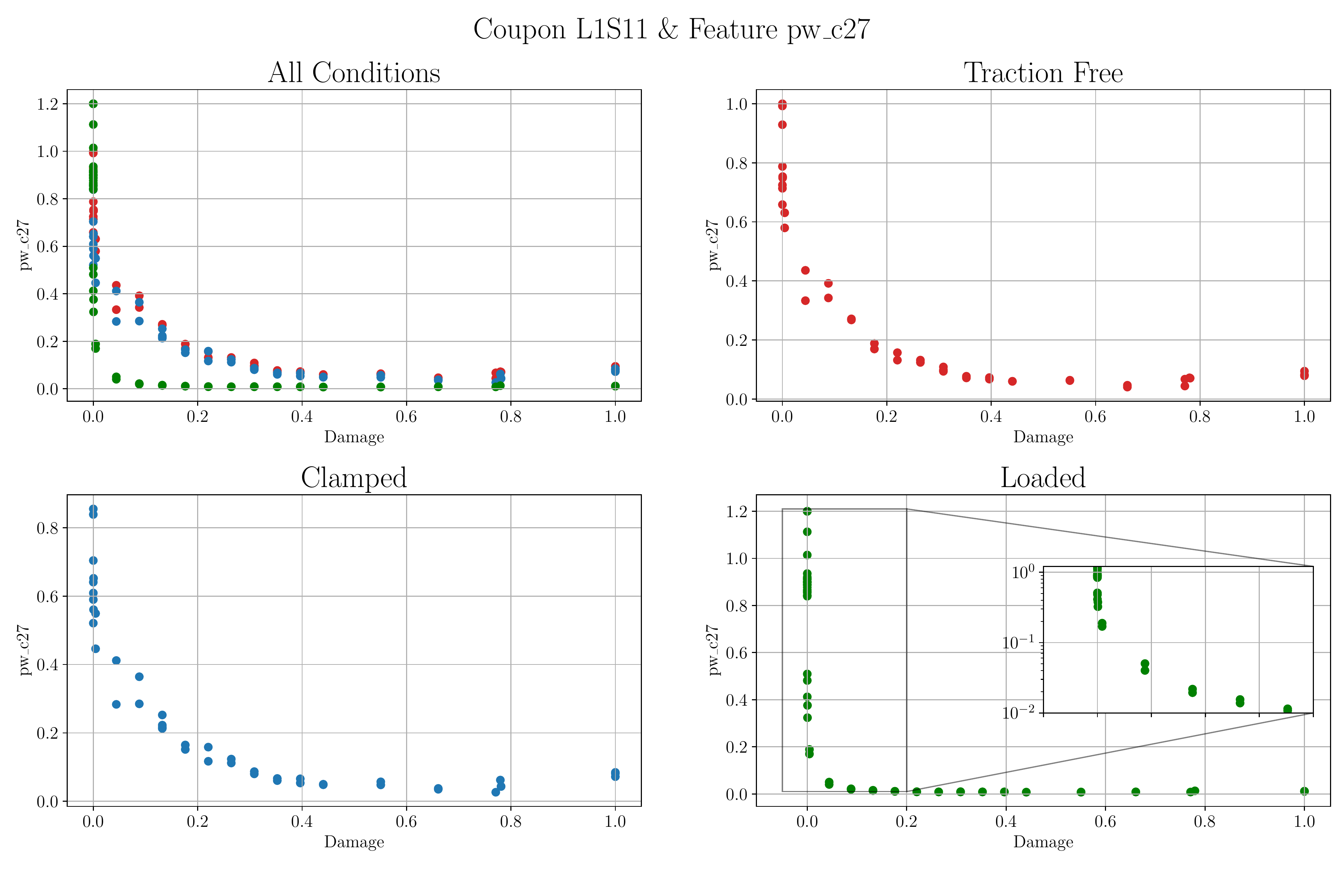}
    \caption{Feature pw\_c27 against damage index in coupon L1S11 for all conditions and by each condition.}
    \label{p27}
\end{figure}

As can be seen, a clear descending tendency is observed as the damage index increases. However, the tendency is better defined when the data points are separated by condition. This indicates that including the condition as a categorical feature will help the algorithms perform better. Note that the drop in power ratio becomes more drastic when measurements are made with the specimen loaded (bottom right plot in \figurename~\ref{p27}). Nevertheless, the decrease in power ratio with the damage is not a sudden fall but a drop at a faster rate. To illustrate this point, a zoom with a logarithmic scale in power ratio has been highlighted. Additionally, similar tendencies are observed in the correlation coefficient features for the same channels.

\begin{figure}[H]
    \centering
    \includegraphics[width = 0.5\textwidth]{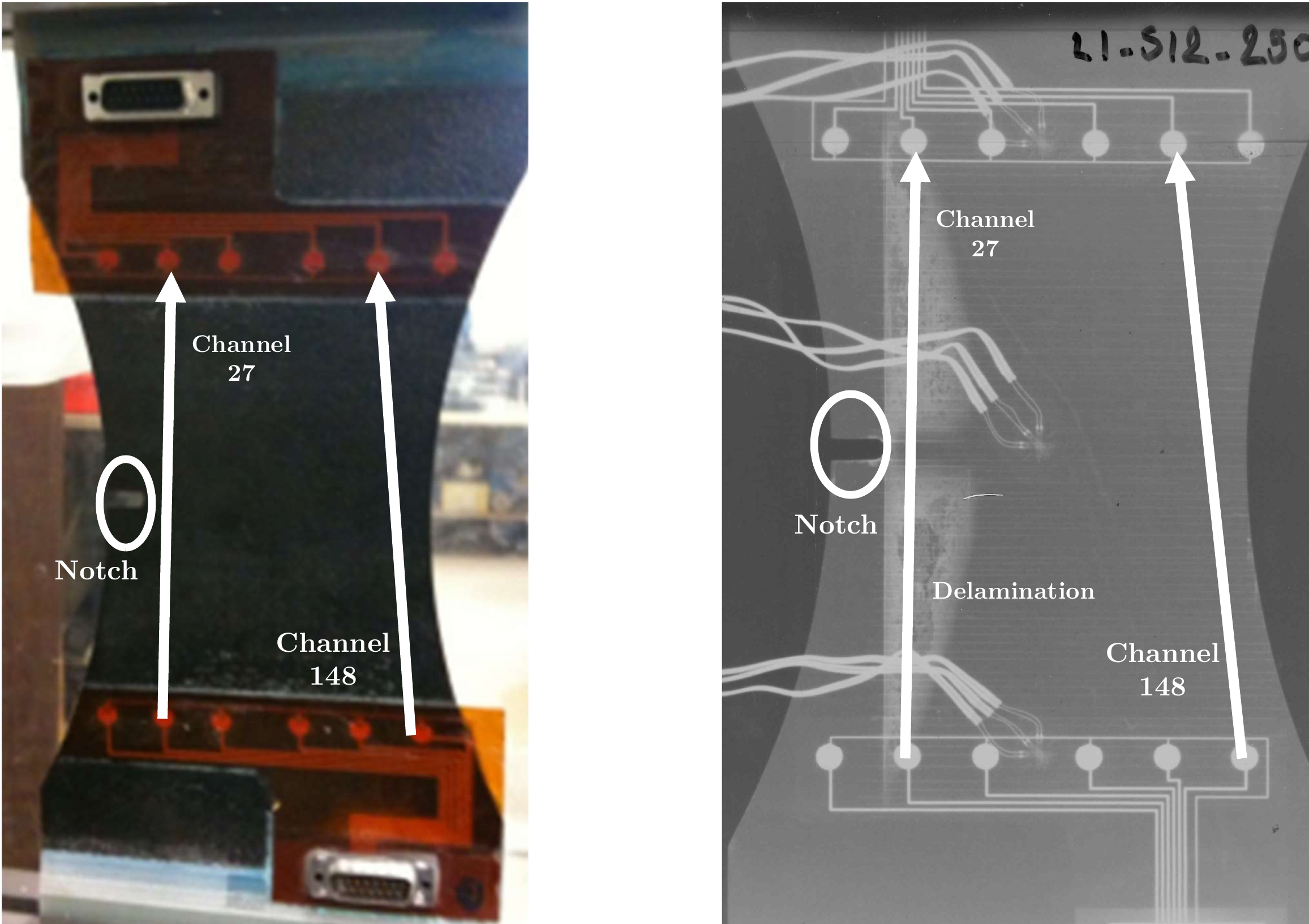}
	\caption{Location of signal paths, notch and delamination. Figure adapted from  \cite{chiachio2013documentation}.}
    \label{paths}
\end{figure}

However, not all channels present this clear patterns. For many channels, the features are just scattered with respect to the damage index, or simply do not vary with it. The explanation for this can be found in the disposal of the sensors in the coupons. Since coupons have a notch to force the delaminations to originate there, the shortest path between some channels will pass straight through the area with great density of defects. Meanwhile, Lamb Waves whose shortest path is away from this area do not encounter defects. Figure \ref{paths} depicts the path of the channel pictured in Figure \ref{p27}, c27, and of c148, whose features do not present any tendency.

As can be seen in the XRay image on the right, the path of channel c27 cuts right through the delaminations whereas that of channel 148 do not encounter noticeable defects. For this reason some features may carry useful information while others may not, and would just introduce noise to the model and cause it to overfit---the model could misallocate resources by learning meaningless variables. However, in a real life scenario there is no certainty of the location of defects. For that reason all channels must be considered and a reliable and systematic way of distinguishing between the information carrying channels and the ones who do not must be proposed. For this regard the \emph{Feature Importance} calculation from Decision Tree algorithms was used as way to rank said features in order of relevance \cite{rogers2005identifying}.

In addition to the power ratio and correlation coefficient of the different channels, more features will be introduced in the model. Since the propagation properties of the Lamb Waves depend on the mechanical properties of the material, the terms of the stiffness matrix will be included to help the algorithm generalize over layups. The load level applied to the loaded condition cases, which in in-service cases may be derived from extensometer measurements, will also be included. As categorical features, the condition and the layup label will also be included. These make a total of 530 features, shown in Table \ref{features}.

\begin{table}[H]
    \centering
    {\small
    \begin{tabularx}{\textwidth}{|l|l|l|X|} 
        \hline
        Feature Type  & Amount & Feature Name & Description  \\
        \hline
        \multirow{4.5}{*}{Power Ratio} & \multirow{4.5}{*}{252} & pw\_c1 & Power ratio for channel 1.  \\
        \hhline{~~--}
        & & \vdots & \vdots  \\
        \hhline{~~--}
        & & pw\_c252 & Power ratio for channel 252.  \\
        \hline
        \multirow{4.5}{*}{\shortstack[c]{Correlation\\ Coefficient}} & \multirow{4.5}{*}{252} & cc\_c1 & Correlation coefficient for channel 1. \\
        \hhline{~~--}
        & & \vdots & \vdots  \\
        \hhline{~~--}
        & & cc\_c252 & Correlation coefficient for channel 252.  \\
        \hline

        \multirow{4.5}{*}{\shortstack[c]{Stiffness\\Matrix\\Terms}} & \multirow{4.5}{*}{18} & A\_11 & Term $A_{11}$ of laminate's stiffness matrix.   \\
        \hhline{~~--}
        & & \vdots & \vdots  \\
        \hhline{~~--}
        & & D\_66 & Term $D_{66}$ of laminate's stiffness matrix.   \\
        \hline

        \multirow{4}{*}{\shortstack[c]{Condition}} & \multirow{4}{*}{4} & condition\_0 & Baseline condition. \\
        \hhline{~~--}
        & & condition\_1 & Clamped condition.  \\
        \hhline{~~--}
        & & condition\_2 & Traction Free condition.  \\
        \hhline{~~--}
        & & condition\_3 & Loaded condition. \\
        \hline
        
        \multirow{3}{*}{\shortstack[c]{Layup}} & \multirow{3}{*}{3} & layup\_1 & Layup 1. \\
        \hhline{~~--}
        & & layup\_2 & Layup 2.   \\
        \hhline{~~--}
       & & layup\_3 & Layup 3.\\
        \hline
        
        Load & 1 & load & Load, only for ``loaded'' cases. \\
        \hline
        \hline
        
        Damage Index& 1 & D & Target of the model.\\
        \hline

    \end{tabularx}}
    \caption{Features and target fed to the models. }
    \label{features}
\end{table}

\subsection{Machine Learning and Hypertuning}

Before feeding the features into the models and obtaining the results, a hypertuning process is performed. The algorithms to hypertune are two FC Neural Networks, one with the first 15 features provided by the feature importance as input and the other, with the amount of features as part of the design space. Also, two Gradient Boost algorithms, LightGBM \cite{ke2017lightgbm} and XGBoost \cite{Chen:2016:XST:2939672.2939785}, as well as Weighted k-Nearest Neighbors \cite{dudani1976distance} are considered. For the first four, the training would be performed with learning rate decay and early stopping, therefore neither the learning rate nor the epochs are part of the design space.

For LightGBM the maximum number of leaves, the maximum depth of each tree and the standard regularization L2 i.e. ridge regularization (regularizing the squared sum of leaf values to avoid a tree growth that leads to overfiting) are considered. For XGBoost, the maximum depth along with two regularization parameters: L2 and gamma---another regularization parameter acting as a penalty for deep nodes. For both Neural Networks, the design space consists of the number of layers, the number of neurons per layer, the batch size---expressed as a fraction of the total amount of data---and the regularization L2. For the second Neural Network, the number of features is also considered. Last, for Weighted k-Nearest Neighbors, the number of neighbors is the hyperparameter to tune. Aditionally, different methods of data scaling are included in the Machine Learning pipeline as hyperparameters. S stands for Standard Scaler (standardization), P for Power Transformer (Yeo-Johnson transformation, \cite{yeo2000new}) and QU and QN for Quantile Transformer with Uniform and Normal output distributions respectively, from the Python library \texttt{sklearn} \cite{scikit-learn}. The design space is shown in Table \ref{ds}.

\begin{table}[H]
    \centering
    {\small
    \begin{tabularx}{\textwidth}{|X|X|X|X|X|X|} 
        \hline
        \multicolumn{5}{|c|}{LightGBM}\\
        \hline
        Hyperparam.& No. Leaves & Max. Depth & L2 & Scaler\\
        \hline
        Range & 5-50 & 10-100 & 0-10 & S,P,QU,QN\\
        \hline
    \end{tabularx}}
\end{table}

\begin{table}[H]
    \centering
    {\small
    \begin{tabularx}{\textwidth}{|X|X|X|X|X|} 
        \hline
        \multicolumn{5}{|c|}{XGBoost}\\
        \hline
        Hyperparam. & Max. Depth & L2 & Gamma & Scaler\\
        \hline
        Range & 3-50 & 0-10 & 0-10 & S,P,QU,QN \\
        \hline
    \end{tabularx}}
\end{table}

\begin{table}[H]
    \centering
    {\small
    \begin{tabularx}{\textwidth}{|X|X|X|X|X|X|} 
        \hline
        \multicolumn{6}{|c|}{Neural Network (15 features)}\\
        \hline
        Hyperparam.& Layers & Neurons & Batch & L2 & Scaler\\
        \hline
        Range & 3-8 & 50-500  & 1-5 & 0-0.1 & S,P,QU,QN\\
        \hline
    \end{tabularx}}
\end{table}

\begin{table}[H]
    \centering
    {\small
    \begin{tabularx}{\textwidth}{| l | *{5}{X |} l |} 
        \hline
        \multicolumn{7}{|c|}{Neural Network ($n$ features)}\\
        \hline
        Hyperparam.& Layers & Neurons & Batch & L2 & Features & Scaler\\
        \hline
       Range & 3-8 & 50-500  & 1-5 & 0-0.1 & 1-530 & S,P,QU,QN\\
        \hline
    \end{tabularx}}
\end{table}

\begin{table}[H]
    \centering
    {\small
    \begin{tabularx}{\textwidth}{|X|X|X|} 
        \hline
        \multicolumn{3}{|c|}{Weighted k-Nearest Neighbors}\\
        \hline
        Hyperparam.& No. Neighbors & Scaler \\
        \hline
        Range & 1-80 & S,P,QU,QN   \\
        \hline
    \end{tabularx}}
    \caption{Hypertuning design space for the used Machine Learning models.}
    \label{ds}
\end{table}

After the proper training, cross validation and test splitting has been made, the hypertuning process is carried out using the algorithm \emph{BOHB} \cite{falkner2018bohb}. Notice that no hypertuning is done over the Maximum Entropy algorithm since all the parameters involved in the model are adjusted during the prediction loop (i.e. no hyperparameter has to be fixed beforehand). This is an advantage for autonomous, self-supervised SHM prediction applications.
\section{Numerical Results}
\label{Sec:results}

\begin{figure}[H]
    \centering
    \includegraphics[width =  0.42\textwidth]{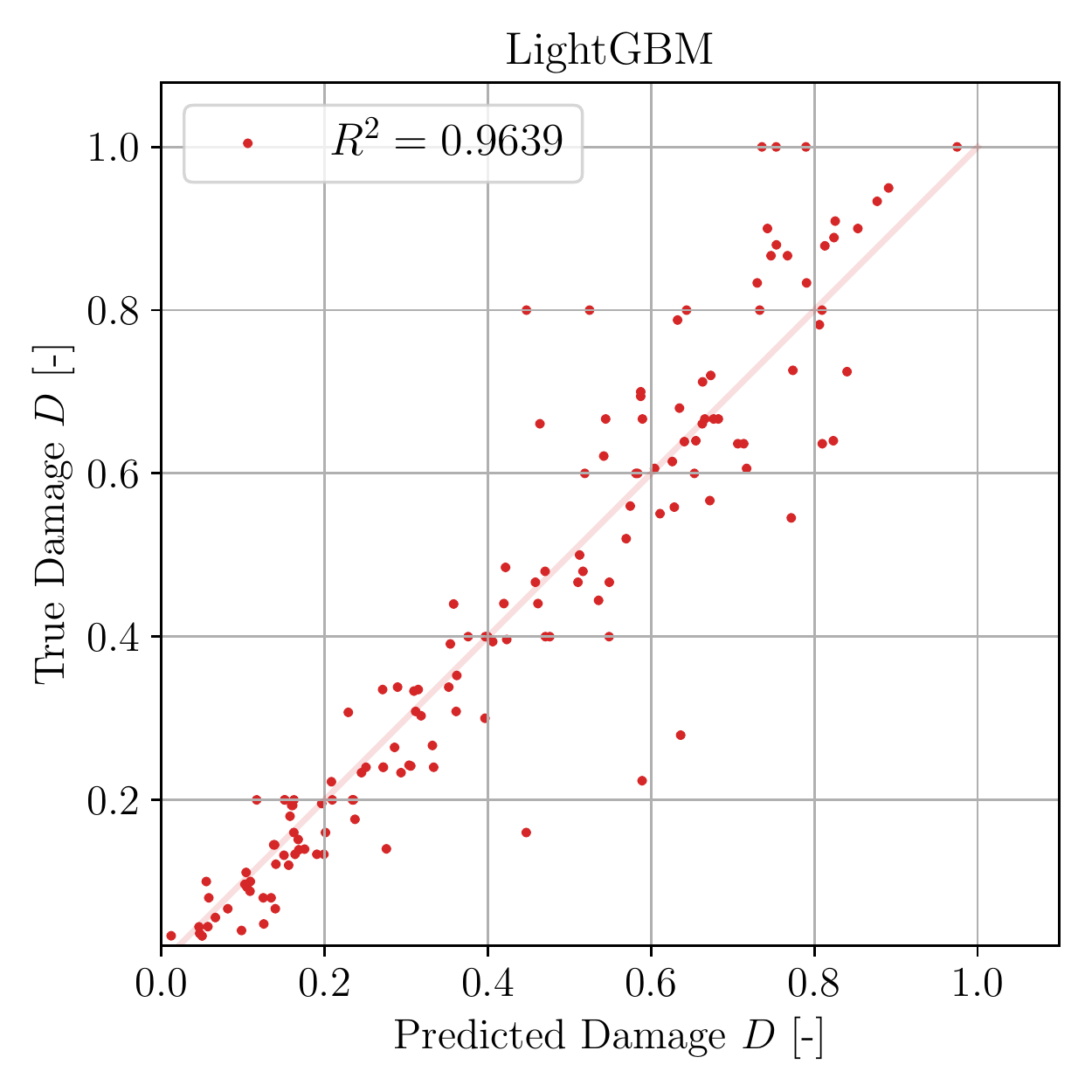}
    \includegraphics[width =  0.42\textwidth]{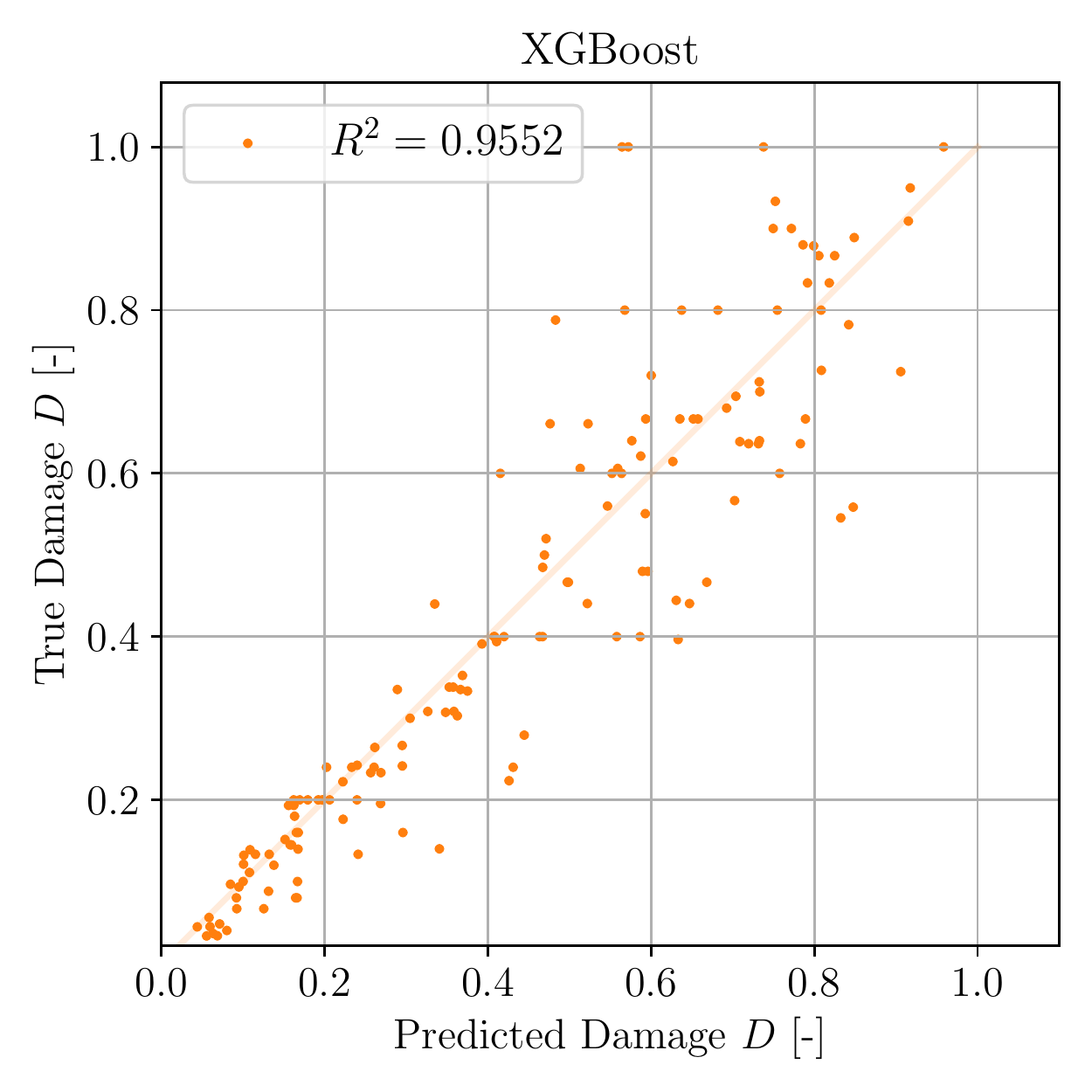}
    \caption{Comparison between the results of Gradient Boost.}
    \label{RES_AL_1}
\end{figure}

\begin{figure}[H]
    \centering
    \includegraphics[width =  0.42\textwidth]{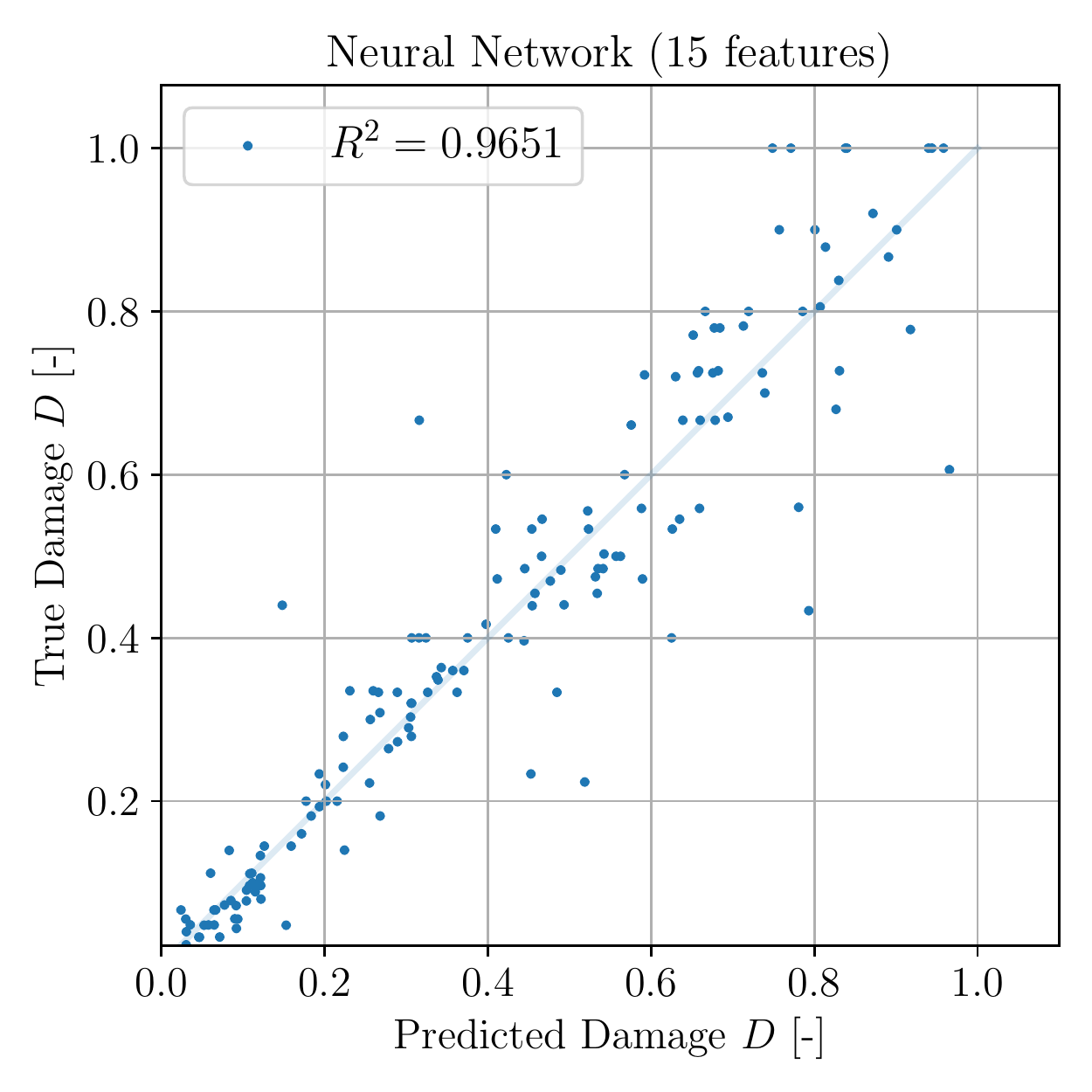}
    \includegraphics[width =  0.42\textwidth]{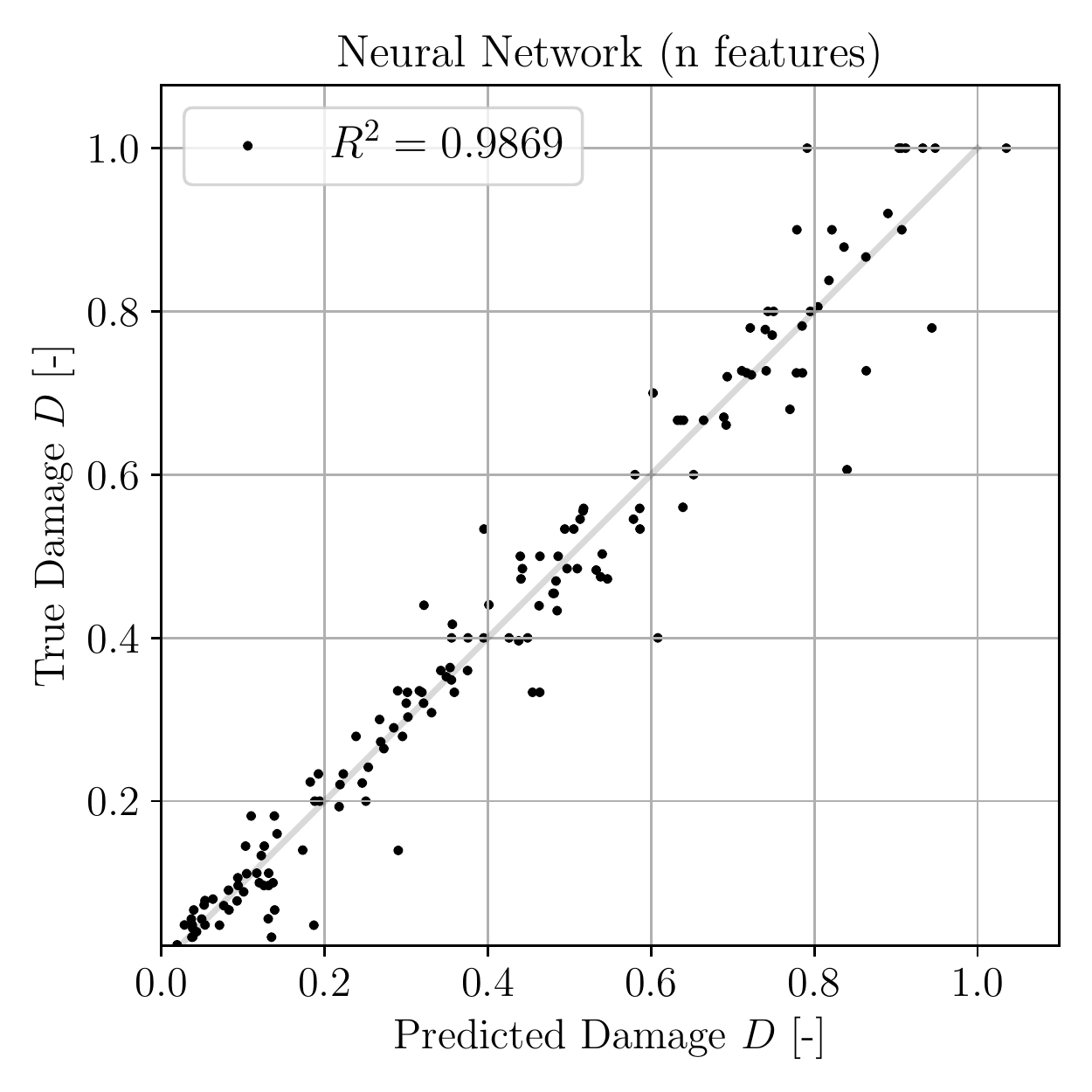}
    \caption{Comparison between the results of Neural Networks.}
    \label{RES_AL_2}
\end{figure}

\begin{figure}[H]
    \centering
    \includegraphics[width =  0.42\textwidth]{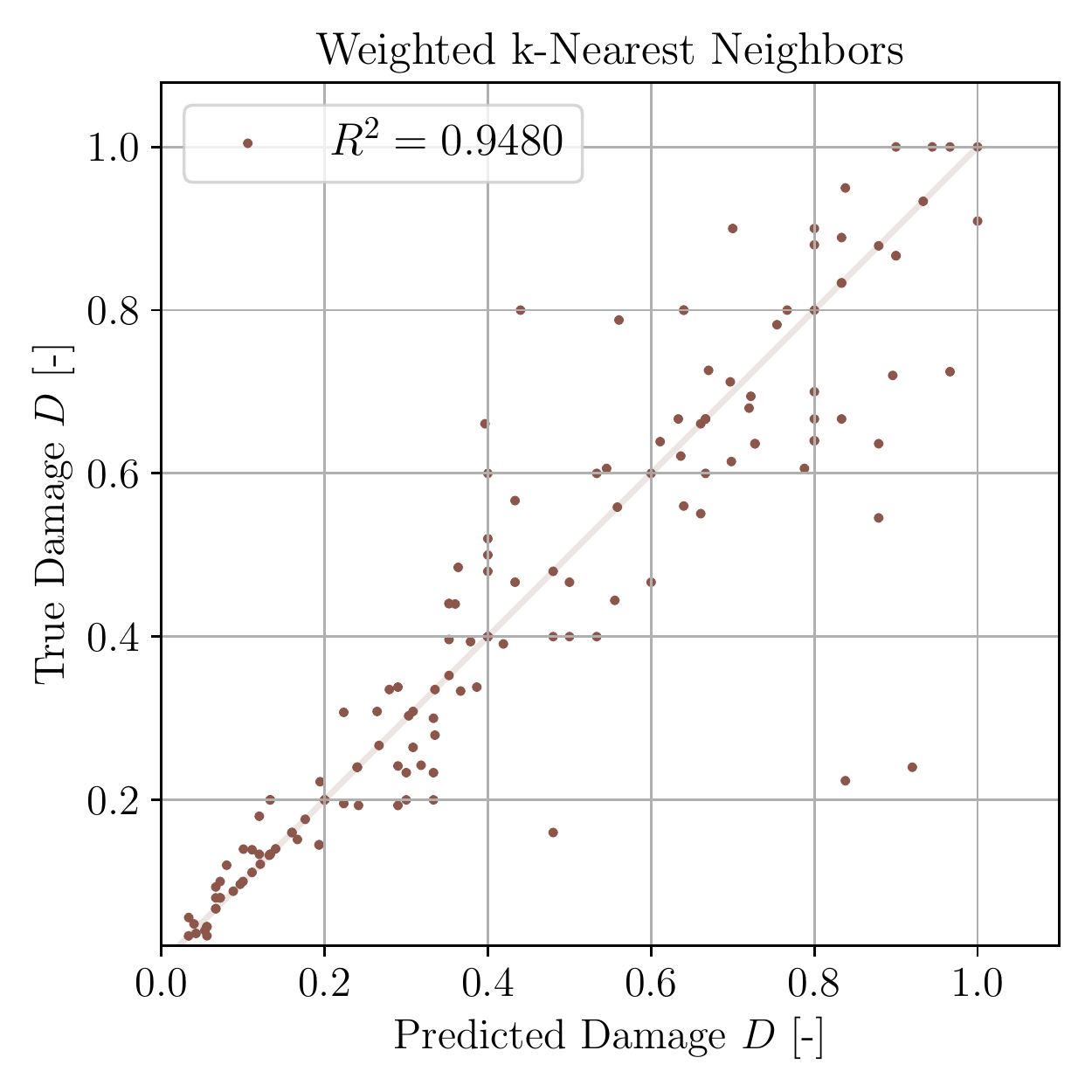}
    \includegraphics[width =  0.42\textwidth]{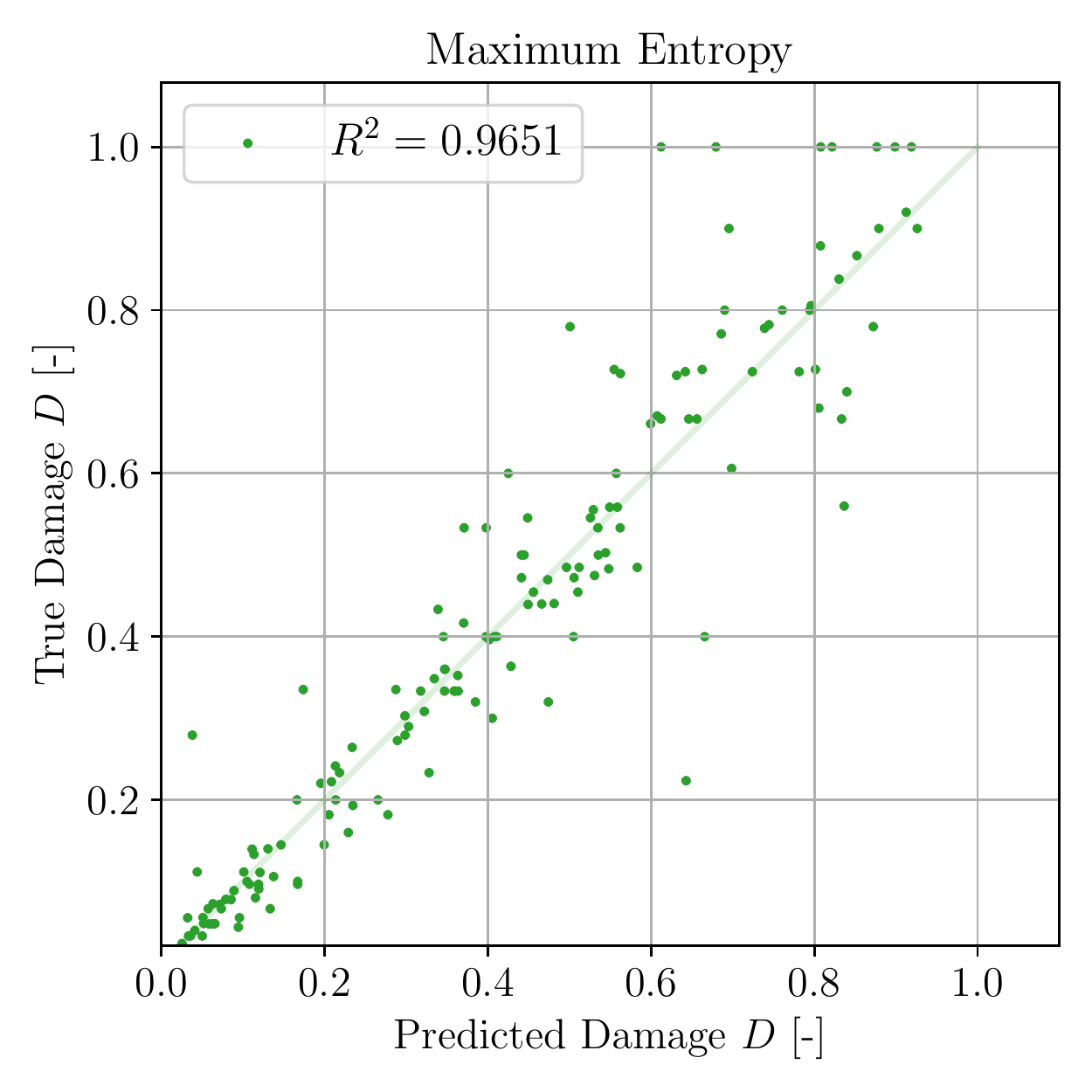}
    \caption{Comparison between the results of k-Nearest Neighbors and Maximum Entropy.}
    \label{RES_AL_3}
\end{figure}

Once the hypertuning is completed, the results obtained are used to fit all mentioned models and evaluate their performance over a test set through the following metric: the coefficient of determination $R^2$. Figures \ref{RES_AL_1}, \ref{RES_AL_2} and \ref{RES_AL_3} depict the true and predicted value over the test set, along with the $R^2$ score of each of them.

For a better comparison, the $R^2$ scores obtained in Figures \ref{RES_AL_1}, \ref{RES_AL_2} and \ref{RES_AL_3} are displayed together in Table \ref{COMP_AL}, along with the Mean Squared Error (MSE) value.

\begin{table}[H]
    \centering
    {\small
    \begin{tabularx}{\textwidth}{|X|X|X|X|X|X|X|X|} 
        \hline
        Algorithm & LightGBM & XGBoost & NN (15) & NN (n) & WKNN & ME\\
        \hline
        $R^2$ & 0.9639 & 0.9552  & 0.9651  & 0.9869  & 0.9480  & 0.9651 \\ 
        \hline
        MSE & 6.580e-3  & 8.127e-3 & 6.337e-3 & 2.4206e-3 & 9.949e-3 & 6.371e-3 \\ 
        \hline
    \end{tabularx}}
    \caption{$R^2$ and MSE scores.}
    \label{COMP_AL}
\end{table}

Figure \ref{RES_AL_2} and Table \ref{COMP_AL} show that the Neural Network with the hypertuned number of features manages to deliver a score close to $R^2 = 0.99$. The Neural Network with 15 features follows it, scoring close to $R^2 = 0.97$. Then comes the Maximum Entropy algorithm achieving over $R^2 = 0.96$, similar to that of LightGBM. XGBoost performs worst, not reaching $R^2 = 0.96$ and, below $R^2 = 0.95$, Weighted k-Nearest Neighbors. As can be seen, the Maximum Entropy algorithm scores significantly better than other similar k-Nearest Neighbors algorithms, and slightly superior than Gradient Boost algorithms.

\begin{figure}[H]
    \centering
    \includegraphics[width = \textwidth]{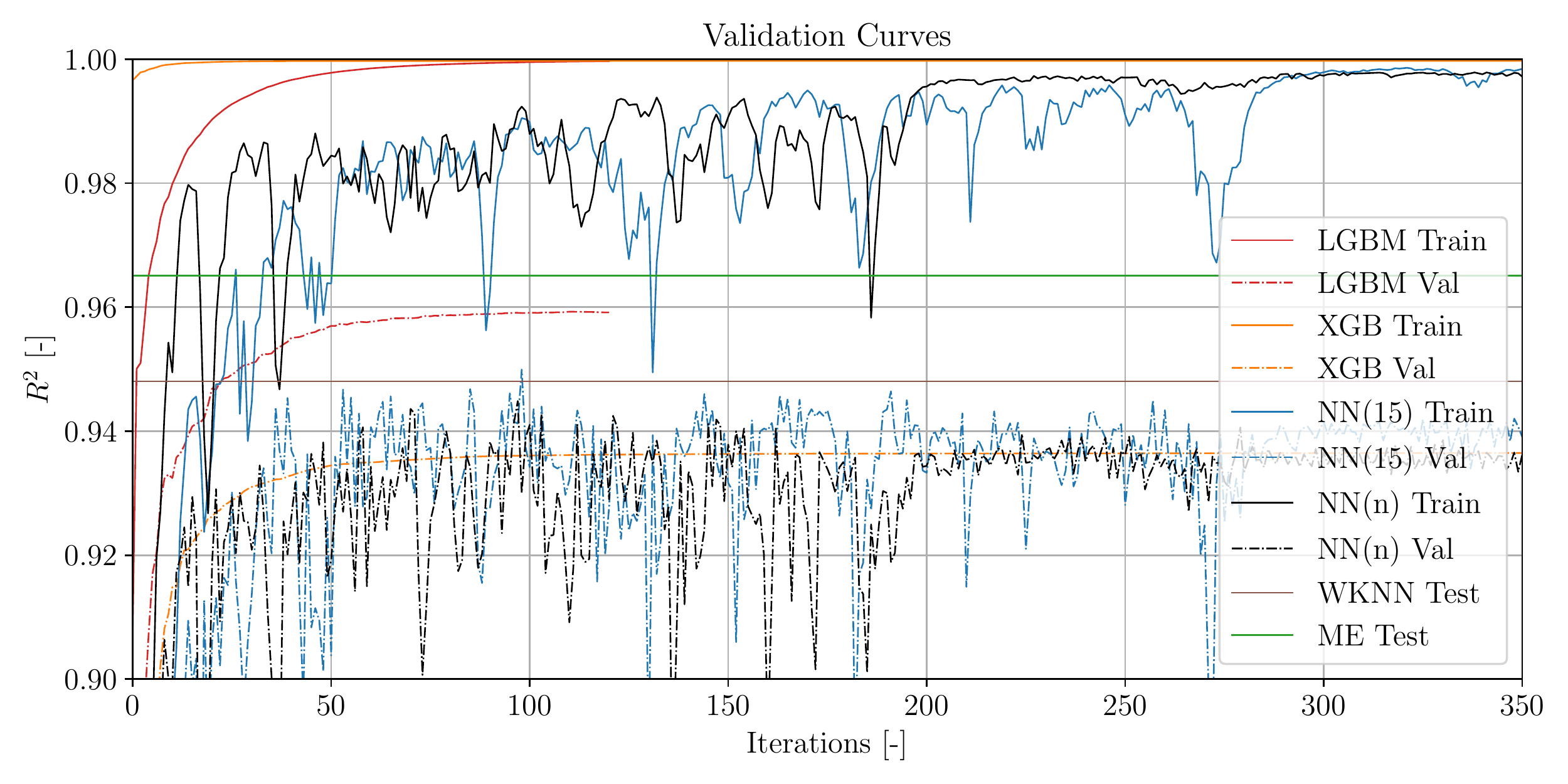}
    \caption{Validation curves.}
    \label{LR_AL}
\end{figure}

The whole analysis will be complete after looking at the validation curves in Figure \ref{LR_AL}, where the learning curves are displayed. 

In Figure \ref{LR_AL} the validation curves over train and validation sets are plotted for the trainable models. For those which do not train, the score over the test set is plotted as an horizontal line for reference. All trainable model curves present great overfitting. In all of them the training score is significantly higher than that in the validation and test sets. This could imply that the model is not regularized sufficiently, but also---given the robustness of BOHB and the fact that it happens across all models---could be due to a too small set of data: it only contains 1492 examples. The training errors for trainable models achieve high scores, but both validation and test errors keep lower than 0.97, except for the test error of one Neural Network.

This suggests that the potential best score for the validation and test set is, at least, that of the current training sets, significantly over $R^2 = 0.99$. Given enough data, from a non skewed distribution and an appropriate regularization to prevent overfitting and the accuracy of the models could reach said scores. Weighted k-Nearest Neighbors and the Maximum Entropy algorithm would also benefit from having a more densely populated input space, and their scores might improve as well.

Finally, the elapsed computational times of all algorithm are presented in Table \ref{TIME_SL}. Examples were run in an Intel(R) Core(TM) i7-6700K CPU @ 4.00GHz (8 CPUs), with 16\,GB RAM.

\begin{table}[H]
    \centering
    \begin{tabularx}{\textwidth}{|c|X|X|X|X|X|X|} 
        \hline
        \multicolumn{7}{|c|}{Computation Time [s]}\\
        \hline
        & LGBM & XGB & NN (15) & NN (n)& WKNN & ME\\
        \hline
        Hypertunning & 1.3e4   & 3.9e4  & 2.8e4  & 4.6e4  & 1.0e4  & 0 \\
        \hline
        Fit & 2.5e0 & 1.3e2  & 1.6e2 & 7.8e1 & 5.0e-3 & 2.0e-3 \\
        \hline
        Prediction & 9.0e-3  & 1.5e-2 & 1.7e-1 & 2.6e-1 & 1.1e-2 & 1.1e2 \\
        \hline
        \textbf{Total} & \textbf{1.3e4}  & \textbf{3.9e4}  &  \textbf{2.8e4} & \textbf{4.6e4} & \textbf{1.0e4} & \textbf{1.1e2} \\
        \hline
    \end{tabularx}
    \caption{Elapsed computational time of all algorithms.}
    \label{TIME_SL}
\end{table}

As it can be seen, the fit and training times of the Maximum Entropy algorithm are comparable to those of Neural Networks, but larger than that of LightGBM and Weighted k-Nearest Neighbor. However, when taking into account the time needed to hypertune the models, which is orders of magnitude greater than the time needed to train and predict, the Maximum Entropy algorithm clearly comes on top. Since most hypertuning algorithms like BOHB require multiple evaluations of the models, this is usually the most time consuming step. Moreover, in \tablename~\ref{TIME_SL} we do not include the time required by the analyst in the hypertuning of these models. This process is skipped entirely with the Maximum Entropy algorithm. 

Additionally, when testing the models for smaller datasets, the total time for the Maximum Entropy algorithm decreases significantly, getting smaller than that of the Neural Network. Since the current implementation of the model predicts one point at a time, it is logical for the total time to be highly sensitive to the number of points to predict. Also, it is important to remark that the first five algorithm implementations are very optimized professional-level Python functions from \texttt{keras}, \texttt{scikit-learn}, \texttt{lightgbm} and \texttt{xgboost}, whereas the Maximum Entropy implementation is an initial testing approach developed for this work. With a detailed benchmark analysis of the program, it would be possible to optimize it to computational times comparable to those of k-Nearest Neighbors algorithms, since they are the most similar ones and the number of operations is similar as well.
\section{Concluding Remarks}
\label{Sec:conclusions}

It has been proven that the proposed Machine Learning workflow is able to arrive to promising results in the determination of the Palgrem-Miner's damage index. By extracting features from the Lamb waves with respect to the baseline measurement that are able to capture the defects growth, Machine Learning algorithms might learn complicated or hard-to-notice patterns and tendencies. Then, they are exploited in order to provide useful information about the health state of the coupon, bypassing the need for a complex analytical model. The accuracy achieved in this project is a good first step towards achieving a reliable and accurate model. To further improve it, increasing the dataset with more measurements and a wider spectrum of layups would be the first approach to outperform the current accuracy.

Regarding the Maximum Entropy algorithm, a successful first implementation has been developed. In terms of accuracy, it has proven to be systemically better than the Weighted k-Nearest Neighbors with the hypertuned number of neighbors. Therefore, it may be one of the most accurate algorithms in the Nearest Neighbors' family, and even be comparable in some cases to more complex algorithms. When it comes to computation times, the current implementation behaves at a certain accuracy level with small sets of points to predict, but it rapidly increases with its size. More sophisticated techniques like vectorized implementations or gradient descent approaches to maximize the mean entropy could be developed to improve efficiency and lower the computational time. Nonetheless, it has the great benefit of not needing hypertuning, which significantly decreases the time needed to set up the model. This could prove particularly useful in applications where the data base is constantly being updated or in \emph{online learning}, i.e. applications where there is a continuous flow of new incoming data. Stochastic approaches need to constantly update the parameters of the models, plus the changing conditions of the database might turn the hypertuned configuration obsolete. With the Maximum Entropy algorithm this would not happen and it would not need to be retrained nor updated, so a constant monitoring of the data could be achieved. Also, the no need for user-set hyperparameters could be useful in stand-alone SHM applications.

\section{Acknowledgements}
\noindent\begin{minipage}[c]{.15\linewidth}
\includegraphics[width=\linewidth]{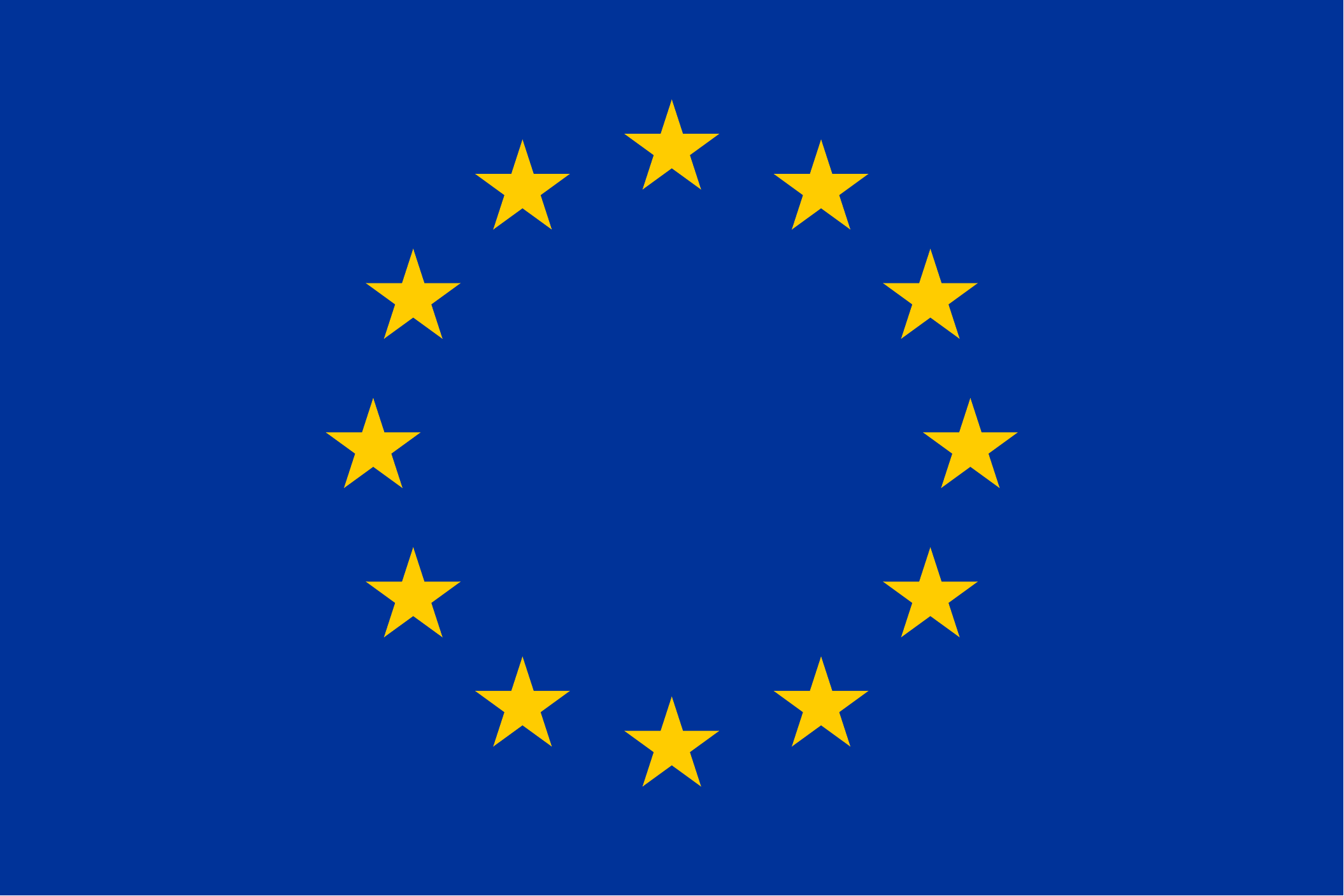}
\end{minipage}\hfill
\begin{minipage}[c]{.8\linewidth}
This project has received funding from the European Union's Horizon 2020 research and innovation programme under the Marie Skłodowska-Curie Grant Agreement No. 101007815.
\end{minipage}

\section{Author contributions}
\noindent\textbf{Miguel Díaz-Lago:} Data Curation, Software, Visualization, Writing-Original draft preparation. \textbf{Ismael Ben-Yelun:} Investigation, Formal analysis, Methodology, Writing-Review \& Editing. \textbf{Luis} \textbf{Saucedo-Mora:} Conceptualization, Software, Supervision. 
\textbf{Miguel Ángel Sanz:} Resources. \textbf{Ricardo Callado:} Resources. \textbf{Francisco Javier Montáns:} Supervision, Project administration, Writing-Review \& Editing.

\appendix
\section{Implementation}\label{Apendx:implementation}

\setcounter{figure}{0}
\setcounter{table}{0}
\renewcommand{\appendixname}{\!}

The proposed implementation is explained hereunder. It calculates the prediction of a single point at a time. For multiple points prediction, the function containing the algorithms is called in parallel, once for every point to predict. Algorithm~\ref{maxentalg} depicts the pseudocode of the algorithm. 

\begin{algorithm}[H]
    \caption{Algorithm Implementation for a Single Point}
    \begin{algorithmic}
        \State Input: $X$, $Y$, $\mathrm{b}$, \texttt{params}
        \State $\mathrm{K_{\mathrm{mat}}} = [X^T|1]$
        \State $b = [b|1]$
        \State $\mathrm{Minconvex} = \mathrm{False}$
        \State $\mathrm{Convergence} = \mathrm{False}$
        \State $\mathrm{error\_old} = q_1$ \Comment{\emph{set by the user}}
        \While{not Minconvex}
            \State $C_1 = \lbrace X^{(i)}:\mathrm{RBF}^{(i)}(\mathrm{h_{filter}})> \mathrm{threshold\_filter} \rbrace$ 
            \For{$h = h_j$}
                \State $C_{2j} = \lbrace X^{(i)} \in C_1:\mathrm{RBF}^{(i)}(h_j)> \mathrm{threshold\_entropy}\rbrace$
                \State $\mathrm{h\_data}_j =  - \frac{1}{m_C} \sum_{i \in C_{2j}} \mathrm{RBF}^{(i)}(h_j) \log \mathrm{RBF}^{(i)}(h_j) $
            \EndFor
            \State $h^* = h_j:\mathrm{h\_data}_j = \max \left( \mathrm{h\_data} \right) $
            \State $C_{2} = \lbrace X^{(i)} \in C_1:\mathrm{RBF}^{(i)}(h^*)> \texttt{threshold\_entropy}\rbrace$
            \State $u^{(i)} = \mathrm{RBF}^{(i)}(h^*)$
            \State $ \mathrm{iter} = 0$
            \While{not Convergence}
                \State $ \mathrm{iter}+= 1$
                \State Solve $K_{\mathrm{mat}}u= b$
                \If{$\mathrm{iter} > \texttt{it\_local\_min}$}
                    \State $\mathrm{h_{filter}} += q_2$ \Comment{ \emph{set by the user}}
                    \If{$||\mathrm{error}| + |1 - \sum_i u_i| - \mathrm{error_{old}}| < \texttt{local\_min\_tolerance}$}
                            \State $\mathrm{Minconvex} = \mathrm{True}$
                            \State $\mathrm{Convergence} = \mathrm{True}$
                            \State $\mathrm{error_{old}} = \mathrm{error}$
                    \EndIf
                \EndIf
            \EndWhile
        \EndWhile
        \If{Regression}
            \State $Y_{\mathrm{pred}} = u Y$
        \EndIf
        \If{Classification}
            \State $Y_{\mathrm{pred}} = \mathrm{mode(Y)}$
        \EndIf
        \State Output: $Y_{\mathrm{pred}}$ 
    \end{algorithmic}
    \label{maxentalg}
\end{algorithm}

Now, each step in the algorithm will be explained.

\begin{enumerate}
    \item \textbf{Input.} $X$, $Y$, $b$ and \texttt{params} are provided.  $X \in \mathbb{R}^{m \times n_x}$ is a matrix with the data examples as rows, and $Y \in \mathbb{R}^{m \times n_y}$ their respective images or labels. $b \in \mathbb{R}^{n_x}$ is a vector with the features of the point to predict. Finally, \texttt{params} contains all thresholds and tolerances required. They are detailed in the Table~\ref{params}.
    
    \begin{table}[htb]
        \centering
        {\small
        \begin{tabularx}{\textwidth}{|c|X|c|} 
            \hline
            \texttt{params} & Description & Default\\
            \hline
            \texttt{threshold\_filter} & Threshold to be used during the filter step. & $0.01$ \\
            \hline
            \texttt{threshold\_entropy} & Threshold to be used during the maximization of entropy step. &  $0.01$\\
            \hline
            \texttt{convergence\_tolerance} & Tolerance of the error plus the sum of interpolation weights for exiting the Convergence and Minconvex loops. & $0.01$ \\
            \hline
            \texttt{it\_convergence} & Minimum amount of iterations for exiting the Convergence and Minconvex loops. & $20$\\
            \hline
            \texttt{local\_min\_tolerance} & Tolerance of the difference between error plus the sum of interpolation weights and the error in the previous iteration for exiting the Minconvex loop. & $1e-9$ \\
            \hline
            \texttt{it\_local\_min} & Minimum amount of iterations for exiting the Convergence loop. & $1000$ \\
            \hline
        \end{tabularx}}
        \caption{Parameters in the Maximum Entropy algorithm.}
        \label{params}
    \end{table}
    
    For the default thresholds to work properly all data fed to the algorithm must be previously \textbf{normalized}, with the majority of data examples in the range $[-1,1]$.

\sps
    
    \item \textbf{Impose $\sum_{i} u^{(i)}$.} $\mathrm{Kmat} = [X^T|1]$, this is, $X$ transposed with an appended columns of ones. Likewise, $b$ is appended a one: $b = [b|1]$.

\sps
    
    \item \textbf{Filter.} Here begins the \texttt{Minconvex} loop. For performance reasons, the maximization of entropy will not be done with all points in the sample. An initial filter removes those points away to the point to predict by imposing a thresholds to their RFBs, \texttt{threshold\_filter}. In practice, the points not in the convex are discarded by removing them from $\mathrm{Kmat}$, $Y$ and $b$.

\sps
    
    \item \textbf{Mean Entropy Maximization.} A \emph{brute force} strategy is implemented. For an initial small value of $h_j$, all points are once again passed through a filter. The new convex in this iteration $j$ will be $ C_{2j}$. For the points remaining in the convex, their mean entropy is calculated and stored in $\mathrm{h\_data}_j$. The parameter $h_j$ is iteratively increased and the values of mean entropy stored. Then, the values are sorted and $h^*$ is retrieved.

\sps
        
    \item \textbf{Removal of Values Out of Convex.} Once the optimal $h^*$ has been found, the points are filtered out of the convex once again: the points not in the convex are removed from $\mathrm{Kmat}$, $Y$ and $b$. Additionally, the weights \texttt{u} are initialized as the RFBs: $u^{(i)} = \mathrm{RBF}^{(i)}, i \in C_2$.

\sps

    \item \textbf{Iteration to Calculate Interpolation Weights.} Here begins the \texttt{Convergence} loop. An iteration to calculate the interpolation weights, \texttt{u}, is performed.

\sps
    
    \item \textbf{Error Estimation.} For every iteration of $u$, the $\mathrm{error}$ is estimated as in Equation (\ref{error}). For this purpose $\mathrm{Kmat_{red}}$ and $b_{\mathrm{red}}$ are defined, which are $\mathrm{Kmat}$ and $b$ eliminating their last column and row respectively.
    
\sps    
    
    \item \textbf{Stopping conditions.} If both conditions
    \begin{equation}\label{convergence_stop1}
        |\mathrm{error}| + |1 - \sum_{i} u^{(i)}| < \texttt{convergence\_tolerance},
    \end{equation}

    \begin{equation}\label{convergence_stop2}
        \mathrm{iter} > \texttt{it\_convergence}
    \end{equation}
    
    are met, the convergence requisite is satisfied and a solution would have been found and both loops are exited. Eitherwise, if
    
    \begin{equation}\label{c2.1}
        \mathrm{iter} > \texttt{it\_local\_min},
    \end{equation}
    
    the execution will pass to the last condition. If not, another iteration in the Convergence loop will take place. This assures that a minimum number of iterations for calculating the interpolation weights have been performed before moving on to the last condition. 
    \begin{equation}\label{localmin_stop}
        \left||\mathrm{error}| + |1 - \sum_{i} u^{(i)}\right| - \mathrm{error_{old}}| < \texttt{local\_min\_tolerance}.
    \end{equation}
    where $\mathrm{error_{old}}$ is the error in the previous iteration. If this is met, the algorithm is considered to have reached a local minimum. Both \texttt{Convergence} and \texttt{Minconvex} loop are exited and the prediction is calculated with the interpolation weights. However, if the condition is not met, an iteration in the \texttt{Minconvex} loop will take place. The parameter $h_{\mathrm{filter}}$ will be enlarged so more points are let through the filter. Then, the whole algorithm is repeated until it is exited by either condition in Equations (\ref{convergence_stop1}) and (\ref{convergence_stop2}) or (\ref{localmin_stop}).
    
	\sps    
    
    \item \textbf{Calculate Prediction.} A function is called to differentiate the regression and classification cases. For regression, the images of the points to predict will be calculated as in Equation (\ref{predreg}), whereas for the classification case as in Equation (\ref{predclas}).
    
	\sps    
    
    \item \textbf{Retrieve Solution.} Either just the predictions or the predictions plus additional information is returned. 
\end{enumerate}

\begin{figure}[H]
    \centering
    \includegraphics[width = 0.85\textwidth]{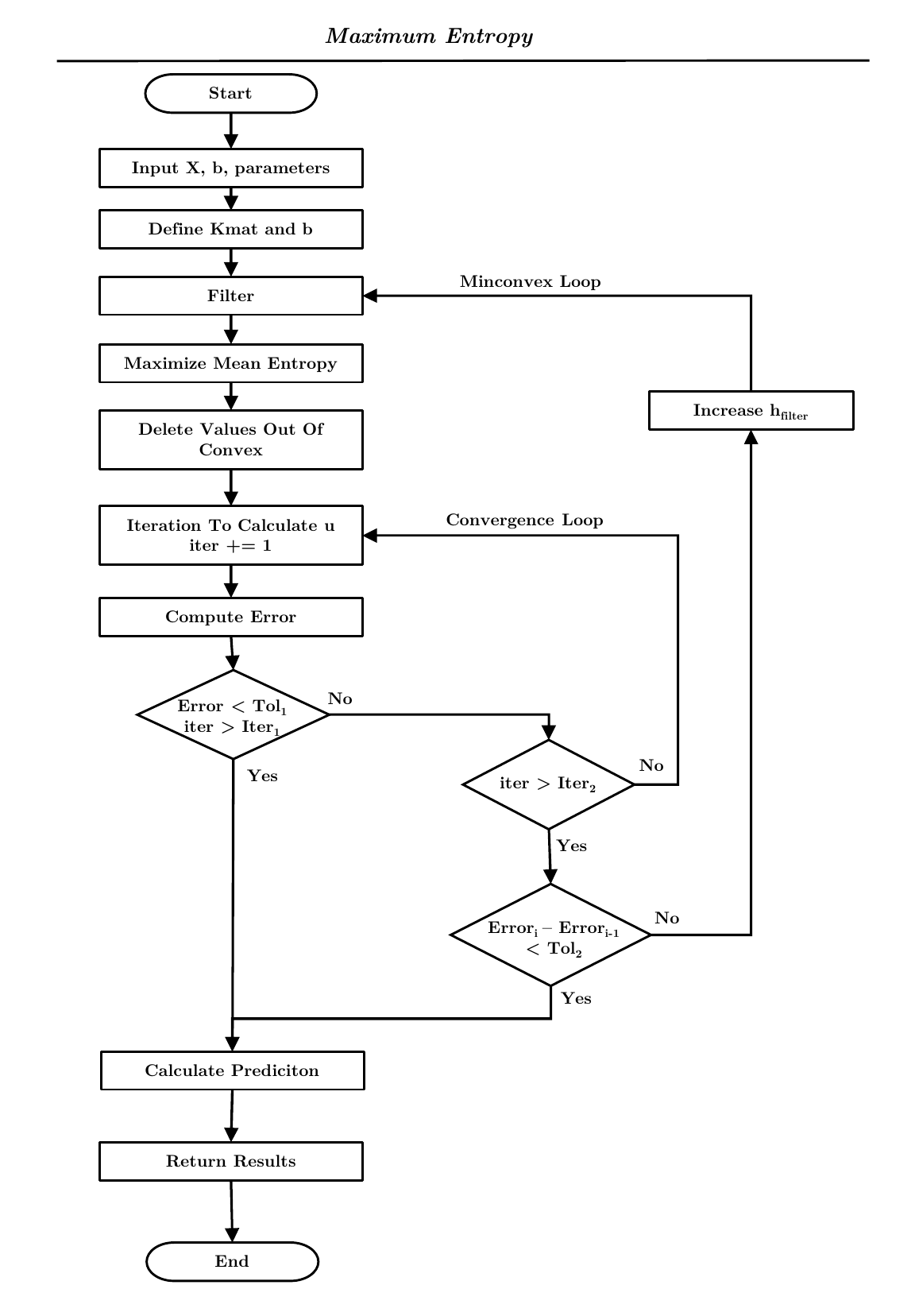}
    \caption{Maximum Entropy algorithm workflow diagram.}
    \label{maxentdiag}
\end{figure}

The workflow diagram of the Maximum Entropy algorithm is depicted in Figure \ref{maxentdiag}.

\addcontentsline{toc}{section}{References}
\bibliographystyle{unsrt-custom}
\bibliography{references.bib}





\end{document}